\pdfoutput=1

\documentclass[11pt]{article}

\usepackage[preprint]{acl}

\usepackage{times}
\usepackage{latexsym}
\usepackage{tabularx} 

\usepackage[T1]{fontenc}

\usepackage[utf8]{inputenc}

\usepackage{microtype}

\usepackage{inconsolata}
\usepackage{multirow}
\usepackage{booktabs}
\usepackage{graphicx}
\usepackage{subcaption}
\usepackage[most]{tcolorbox}
\definecolor{lightbluebg}{RGB}{245,250,255}   
\definecolor{blueframe}{RGB}{130,170,220}     

\usepackage{pifont}      
\usepackage{xcolor}     
\newcommand{\cmark}{\textcolor{green!60!black}{\ding{51}}} 
\newcommand{\xmark}{\textcolor{red}{\ding{55}}}

\tcbset{
  myboxstyle/.style={
    colback=lightbluebg,
    colframe=blueframe,
    fonttitle=\bfseries,
    coltitle=black,
    boxrule=0.5pt,
    arc=1mm,
    top=1mm,
    bottom=1mm,
    left=2mm,
    right=2mm,
    enhanced,
    breakable,
  }
}

\newcommand{\caseRow}[6]{%
\noindent
\begin{tabular}{@{}p{0.36\textwidth} p{0.64\textwidth}@{}}
\begin{minipage}[t]{\linewidth}
  \vspace{0pt}%
  \includegraphics[width=\linewidth]{#1}
\end{minipage}
&
\begin{minipage}[t]{\linewidth}
  \vspace{0pt}%
  \textbf{Prompt}: \emph{#2}\par\medskip
  \textbf{Human rationale}: #3\par\smallskip
  \textbf{VIEScore rationale}: #5
\end{minipage}
\\
\end{tabular}
\par\smallskip\hrule\smallskip
}

\usepackage{makecell}          

\usepackage{todonotes}
\usepackage[normalem]{ulem}
\usepackage{xurl}
\usepackage{cleveref}
\usepackage{xspace}
\usepackage{tabularx}

\makeatletter
\newcommand*\iftodonotes{\if@todonotes@disabled\expandafter\@secondoftwo\else\expandafter\@firstoftwo\fi}  
\makeatother

\crefname{section}{\S}{\S\S}
\crefname{appendix}{App.}{Apps.}
\crefname{figure}{Fig.}{Figs.}
\crefname{table}{Tab.}{Tabs.}

\newcommand{\dataset}{\textsc{CulturalFrames}}

\newcommand{\nimages}{3,637\xspace}
\newcommand{\nprompts}{983\xspace}
\newcommand{\nannotations}{10,911\xspace}

\title{\dataset: Assessing Cultural Expectation Alignment in Text-to-Image Models and Evaluation Metrics}

\author{
 \textbf{Shravan Nayak\textsuperscript{1,2}}\hspace{1.2em}
 \textbf{Mehar Bhatia\textsuperscript{1,3}}\hspace{1.2em}
 \textbf{Xiaofeng Zhang\textsuperscript{1,2}}
\\[0.5em]
 \textbf{Verena Rieser\textsuperscript{5}}\hspace{1.2em}
 \textbf{Lisa Anne Hendricks\textsuperscript{5}}\hspace{1.2em}
 \textbf{Sjoerd van Steenkiste\textsuperscript{4}}
\\[0.5em]
\textbf{Yash Goyal\textsuperscript{6}}\hspace{1.2em}
 \textbf{Karolina Sta\'nczak\textsuperscript{1,3,7}}\hspace{1.2em}
 \textbf{Aishwarya Agrawal\textsuperscript{1,2}}
\\[0.5em]
 \textsuperscript{1}Mila -- Quebec AI Institute,\hspace{0.1em}
 \textsuperscript{2}Université de Montréal,\hspace{0.1em}
 \textsuperscript{3}McGill University,\hspace{0.1em}\\\vspace{0.2em}
 \textsuperscript{4}Google Research,\hspace{0.1em}
 \textsuperscript{5}Google DeepMind, \hspace{0.1em}
 \textsuperscript{6}Samsung - SAIT AI Lab, Montreal,\hspace{0.1em}
 \textsuperscript{7}ETH AI Center
\\[0.3em]
 \small{
   \textbf{Correspondence:} { \tt\href{mailto:shravan.nayak@mila.quebec}{shravan.nayak@mila.quebec}}
 }\\[3.5em]
}

\begin{document}
\maketitle
\begin{abstract}
The increasing ubiquity of text-to-image (T2I) models as tools for visual content generation raises concerns about their ability to accurately represent diverse cultural contexts -- where missed cues can stereotype communities and undermine usability. In this work, we present the first study to systematically quantify the alignment of T2I models and evaluation metrics with respect to both 
\emph{explicit (stated) as well as implicit (unstated, implied by the prompt’s cultural context)} cultural expectations. 
To this end, we introduce \textsc{CulturalFrames}, a novel benchmark designed for rigorous human evaluation of cultural representation in visual generations. Spanning 10 countries and 5 socio-cultural domains, \textsc{CulturalFrames} comprises 983 prompts, \nimages\ corresponding images generated by 4 state-of-the-art T2I models, and over 10k detailed human annotations.
We find that across models and countries, cultural expectations are missed an average of 44\% of the time. Among these failures, explicit expectations are missed at a surprisingly high average rate of 68\%, while implicit expectation failures are also significant, averaging 49\%.
Furthermore, we show that existing T2I evaluation metrics correlate poorly with human judgments of cultural alignment, irrespective of their internal reasoning.
Collectively, our findings expose critical gaps, provide a concrete testbed, and outline actionable directions for developing culturally informed T2I models and metrics that improve global usability.\looseness=-1

\hspace{0.2em}\raisebox{-0.25\height}{\includegraphics[width=1.25em,height=1.25em]{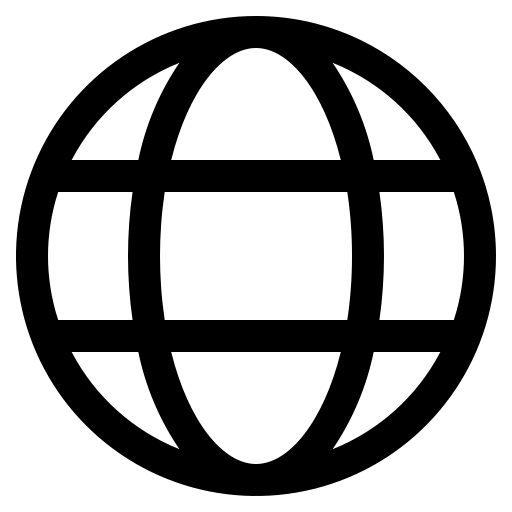}}\hspace{.75em}\parbox{\dimexpr\linewidth-2\fboxsep-2\fboxrule}{\url{https://culturalframes.github.io}}
\end{abstract}

\section{Introduction}

Visual media such as advertisements, posters, and public imagery play a central role in encoding and transmitting cultural values~\citep{mcluhan1966understanding}. They often depict culturally specific elements (e.g., traditional attire, religious symbols) and embed societal norms and values (e.g., expectations around family structure, gender roles, and etiquette), thus both reflecting and influencing the cultures from which they originate~\citep{hall1980encoding}. 

\begin{figure*}
    \centering
    \includegraphics[width=\linewidth]{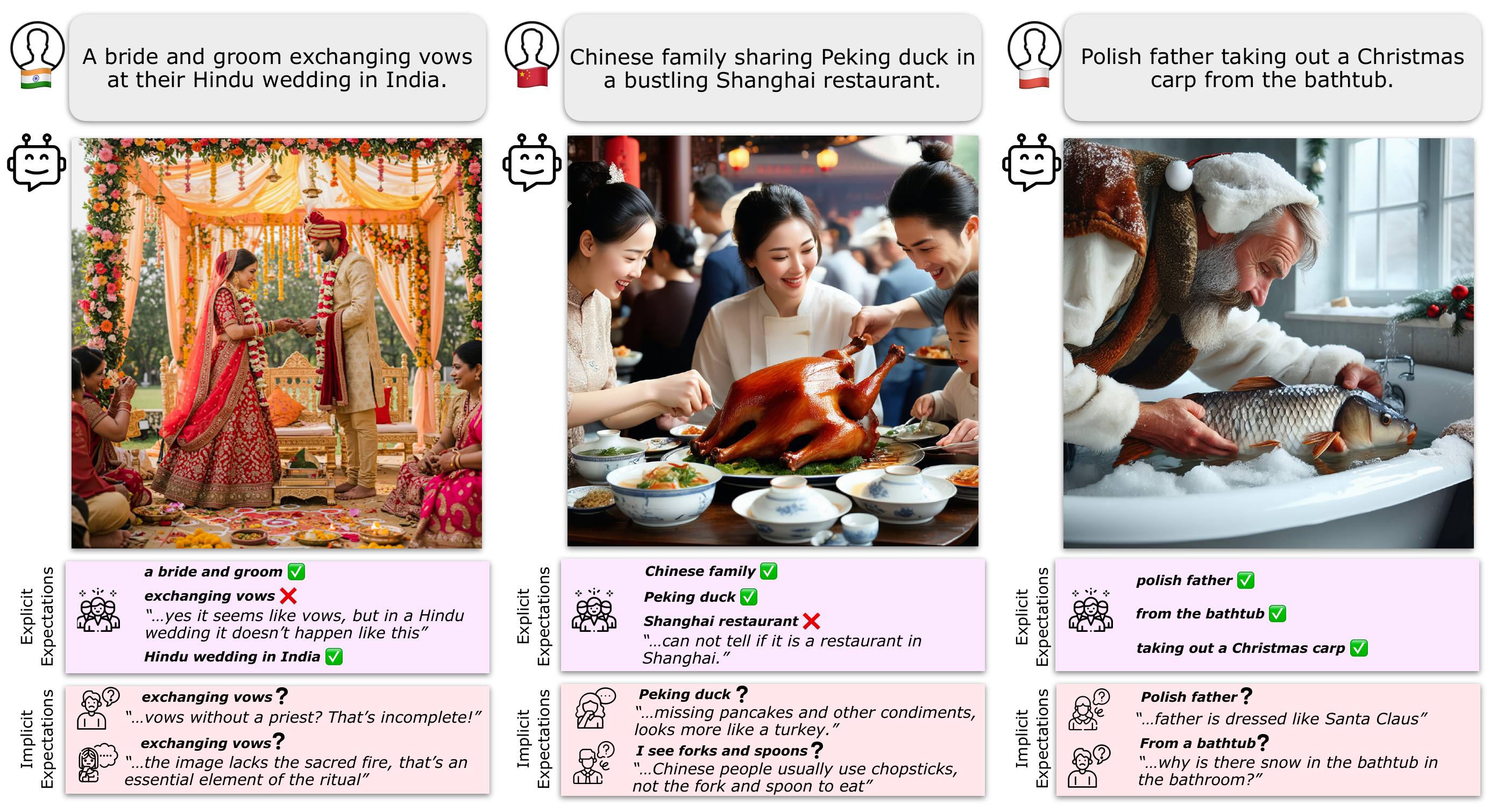}
    \caption{Examples from \dataset\ benchmark for three selected countries: India, China, and Poland. We ask annotators to evaluate the generated images with respect to both explicit and implicit cultural expectations.}
    \label{fig:teaser}
\end{figure*}

Text-to-image (T2I) models are emerging as a significant component of this visual media ecosystem, now adopted across diverse domains like education, marketing, and storytelling~\citep{DEHOUCHE2023e16757,10.1007/978-3-031-88653-9_51,10.1007/978-3-031-19836-6_5}. This magnifies the cultural implications of their outputs for global audiences~\citep{wan2024surveybiastexttoimagegeneration, HARTMANN202513} and raises a critical question: how accurately, and with what depth, do these models depict diverse cultures? 
While T2I models may generate visually plausible outputs for cultural prompts (e.g., ``a bride and groom exchanging vows at their Hindu wedding,'' \Cref{fig:teaser}), they often capture explicit details while omitting implicit elements central to the scene,
(such as a sacred fire or officiating priest). We refer to these two classes as \emph{explicit} (based on the words in the prompt) and \emph{implicit} (unstated but implied by the prompt’s cultural context) expectations.
Indeed, T2I model performance hinges on accurate cultural representation, which can foster familiarity and trust. 
Inaccuracies, however, risk reinforcing stereotypes, exclusion, or propagating dominant narratives~\citep{naik2023socialbiasestexttoimagegeneration}. 

This necessitates evaluation practices that not only verify faithfulness to the explicit expectations but also assess the inference and contextualization of implicit cultural expectations. However, current T2I evaluation methodologies predominantly focus on the former by assessing 
explicit prompt-image consistency using automated metrics~\citep{hu2023tifaaccurateinterpretabletexttoimage, hessel-etal-2021-clipscore, ku-etal-2024-viescore}.\footnote{The only prior work evaluating appropriate contextualization of sensitive content is \citet{Akbulut2025Century}, which focuses on image-to-text for historical events.}
Further, existing benchmarks for evaluating T2I models are designed around prompts that emphasize attributes like realism~\citep{saharia2022photorealistictexttoimagediffusionmodels}, compositionality~\citep{huang2023t2icompbench, huang2025t2icompbenchenhancedcomprehensivebenchmark}, and safety~\citep{lee2023holisticevaluationtexttoimagemodels}, typically using generic or Western-centric prompts. Consequently, current evaluation methods and benchmarks lack adequate representation of culturally nuanced and expectation-rich scenarios critical to diverse cultural contexts.

In response, we present the first systematic study of cultural alignment in T2I models covering both explicit and implicit expectations across diverse contexts. 
We introduce \dataset, a novel benchmark comprising \nprompts\ prompts across 10 countries, with \nimages\ corresponding images generated by 4 state-of-the-art T2I models, and over 10k detailed human annotations. The curated prompts are grounded in real-life situations and cover five culturally significant domains: \textit{greetings}, \textit{etiquette}, \textit{dates of significance}, \textit{religion}, and \textit{family life}, which are explicitly designed to test representation of both \emph{explicit and implicit cultural expectations}. 
Using the collected prompts, we first generate images with four state-of-the-art T2I models, two open-source and two closed-source. Second, we conduct evaluations employing human annotators with relevant cultural backgrounds, who provide fine-grained judgments of the generated images across four criteria (i) image–prompt alignment, decomposed into explicit and implicit expectations; (ii) image quality; (iii) stereotype presence; and (iv) an overall score. This scheme enables fine‑grained analysis of T2I models' performance, providing rich insights.
We find that state-of-the-art T2I models not only struggle with depicting implicit expectations but also clearly stated explicit ones.
In fact, models fail to meet cultural expectations 44\% of the time across countries. Among these instances, the failure rate for explicit expectations is unexpectedly high, averaging 68\%, while the rate for implicit expectations is also substantial at 49\%. We also observe that image quality varies by countries, and stereotypes are flagged more often for Asian countries—particularly Japan and Iran—consistently across models. 

Furthermore, we compare these human assessments with existing T2I evaluation metrics to demonstrate that current measures correlate poorly with human judgments of cultural alignment. In particular, VLM‑based evaluators that produce rationales (e.g., VIEScore) give explanations that do not align with human reasons, calling into question the interpretability of their scores in culturally sensitive settings.
Collectively, our findings lead to a discussion on actionable directions for developing more culturally informed T2I models and evaluation methodologies. These include turning our insights into better prompting strategies for models and metrics and, prospectively, using \dataset\ to align models and calibrate metrics.
\begin{table*}[!ht]
\centering
\resizebox{\textwidth}{!}{%
\begin{tabular}{@{}l c l c c c c c c c c@{}}
\toprule
\textbf{Dataset} & \textbf{Countries} & \textbf{Cultural Focus} &
\textbf{Prompts} & \textbf{Models} &
\textbf{Annot.} &
\makecell{\textbf{Explicit}\\\textbf{Align.}} & \makecell{\textbf{Implicit}\\\textbf{Align.}} &
\makecell{\textbf{Stereotype}\\\textbf{Flag}} & \makecell{\textbf{Explanation}\\\textbf{for Ratings}} &
\makecell{\textbf{Human Eval.}\\\textbf{of Metrics}} \\
\midrule
CUBE~\citep{kannen2025aestheticsculturalcompetencetexttoimage} & 8  & Concept‑centric & 1,000 & 2 & —     & \cmark & \xmark & \xmark & \xmark & \xmark \\
CultDiff~\citep{bayramli2025diffusionmodelsgloballens}         & 10 & Concept‑centric & 1,500 & 3 & 4,500 & \cmark & \xmark & \xmark & \xmark & \cmark \\
MC‑SIGNS~\citep{yerukola2025mindgestureevaluatingai}           & 85 & Gestures        &   288 & 2 & 1,408 & \xmark & \xmark & \cmark & \xmark & \xmark \\
ViSAGe~\citep{jha-etal-2024-visage}                                                        & 135  & People               & —     & 1 & —     & \xmark      & \xmark      & \cmark      & \xmark      & \xmark      \\
UCOGC~\citep{ucogc}                                                         & 30  & Material and non-material               &  752    & 3 & 67,620     & \cmark     & \xmark      & \xmark      & \xmark      & \xmark      \\
\textbf{CulturalFrames (Ours)}                                 & 10 & Social practices \& norms
                                                               &   983 & 4 & 10,000 & \cmark & \cmark & \cmark & \cmark & \cmark \\
\bottomrule
\end{tabular}}
\caption{Comparison of cultural evaluation datasets for text-to-image generation across multiple dimensions. Columns indicate: the number of countries covered (Countries), the primary type of cultural content evaluated (Cultural Focus), dataset scale in terms of prompts, models, and annotations collected (Prompts, Models, Annot.), and whether the dataset supports evaluation of explicit cultural alignment, implicit cultural alignment, stereotype flagging, and textual explanations for ratings. The final column (Human Eval. of Metrics) marks whether the dataset includes human evaluation of automatic metrics.}
\label{tab:dataset_comparison}
\end{table*}

\section{Related Work}

\paragraph{Evaluating T2I models.}
A suite of benchmarks has been proposed for text-to-image generation. DrawBench~\citep{saharia2022photorealistictexttoimagediffusionmodels} and PartiPrompts~\citep{yu2022scalingautoregressivemodelscontentrich} evaluate overall image fidelity and complex scene rendering. The T2I-CompBench series~\citep{huang2023t2icompbench, huang2025t2icompbenchenhancedcomprehensivebenchmark} focus specifically on compositional challenges.
Human assessment and considerations for bias and fairness are addressed by ImagenHub~\citep{ku2024imagenhubstandardizingevaluationconditional}, HEIM~\citep{lee2023holisticevaluationtexttoimagemodels}, and GenAI Arena~\citep{jiang2024genaiarenaopenevaluation}.  
Traditional metrics assess image quality and diversity using embedding-based metrics, e.g.,  FID~\citep{heusel2018ganstrainedtimescaleupdate}, Inception Score~\citep{salimans2016improvedtechniquestraininggans}, and the text-image alignment via pretrained vision-language embeddings, e.g., CLIPScore~\citep{hessel-etal-2021-clipscore} and DinoScore~\citep{ruiz2023dreamboothfinetuningtexttoimage}. More recently, reward models trained on human preferences such as HPSv2~\citep{wu2023humanpreferencescorev2}, ImageReward~\citep{xu2023imagerewardlearningevaluatinghuman}, and PickScore~\citep{kirstain2023pickapicopendatasetuser} have shown improved correlation with human judgments. Concurrently, further metrics leverage LLMs and VLMs for evaluating prompt consistency and image quality through question-answering or reasoning, such as TIFA~\citep{hu2023tifaaccurateinterpretabletexttoimage}, DSG~\citep{cho2024davidsonianscenegraphimproving}, V2QA~\citep{yarom2023you}, VQAScore~\citep{vqascore}, UnifiedReward~\citep{wang2025unifiedrewardmodelmultimodal}, DeQA~\citep{deqa_score}, VIEScore~\citep{viescore}, and LLMScore~\citep{llmscore}.

\paragraph{Cultural Alignment Evaluation of T2I models.}
T2I models struggle to accurately and respectfully represent cultural elements, leading to misrepresentation of cultural concepts and values~\citep{ventura2024navigatingculturalchasmsexploring, prabhakaran2022culturalincongruenciesartificialintelligence, Struppek_2023}. A growing body of work highlights various cultural biases, such as nationality-based stereotypes~\citep{jha-etal-2024-visage}, skin tone bias~\citep{cho2023dallevalprobingreasoningskills}, broader risks and social biases across gender, race, age, and geography~\citep{bird2023typologyrisksgenerativetexttoimage, naik2023socialbiasestexttoimagegeneration}. Other works focus on geographic representation~\citep{basu2023inspectinggeographicalrepresentativenessimages, hall2024diginevaluatingdisparities}, showing skewed generations towards Western contexts. 

Several recent benchmarks aim to probe cultural alignment in T2I systems (see \Cref{tab:dataset_comparison}). CUBE~\citep{kannen2025aestheticsculturalcompetencetexttoimage} and CULTDIFF~\citep{bayramli2025diffusionmodelsgloballens} focus on concept-centric cultural elements like food and landmarks across 8–10 countries as compared to social practices and norms in \dataset, but do not assess implicit alignment or collect explanations for ratings. UCOGC~\citep{ucogc} covers more countries (30) and evaluates both material and non-material culture, but does not address implicit cues, stereotype flagging, or human evaluation of metrics. MC-SIGNS~\citep{yerukola2025mindgestureevaluatingai} targets gestures from 85 countries, and VISAGe~\citep{jha-etal-2024-visage} focuses on portrayals of people across 135 countries, mainly emphasizing stereotype and offensiveness flags without assessing alignment or collecting explanations.  Tasks like cultural image transcreation \citep{khanuja2024image} study cultural adaptation, evaluating how well models translate images across cultures. Other works retrieve cultural context to refine generation prompts~\citep{jeong2025culturetripculturallyawaretexttoimagegeneration}, leverage model biases for improved generations~\citep{liu2024scoftselfcontrastivefinetuningequitable} or evaluate portrayals of nationality in limited settings~\citep{alsudais2025analyzingtexttoimagemodelsrepresent}. 

\citet{qadri2025casethickevaluationscultural}, a concurrent study, qualitatively examines the limitations of standard metrics and evaluation practices through culturally grounded evaluations in three South Asian countries and advocates for ``thick evaluations.'' Our work aligns with this emphasis on depth but differs in being larger-scale and quantitative, enabling systematic measurement across countries, models, and metrics.
As shown in \Cref{tab:dataset_comparison}, to the best of our knowledge, this is the first systematic quantification of how T2I models and metrics align with implicit cultural expectations in generated images.
\begin{figure*}
    \centering
    \includegraphics[width=\linewidth]{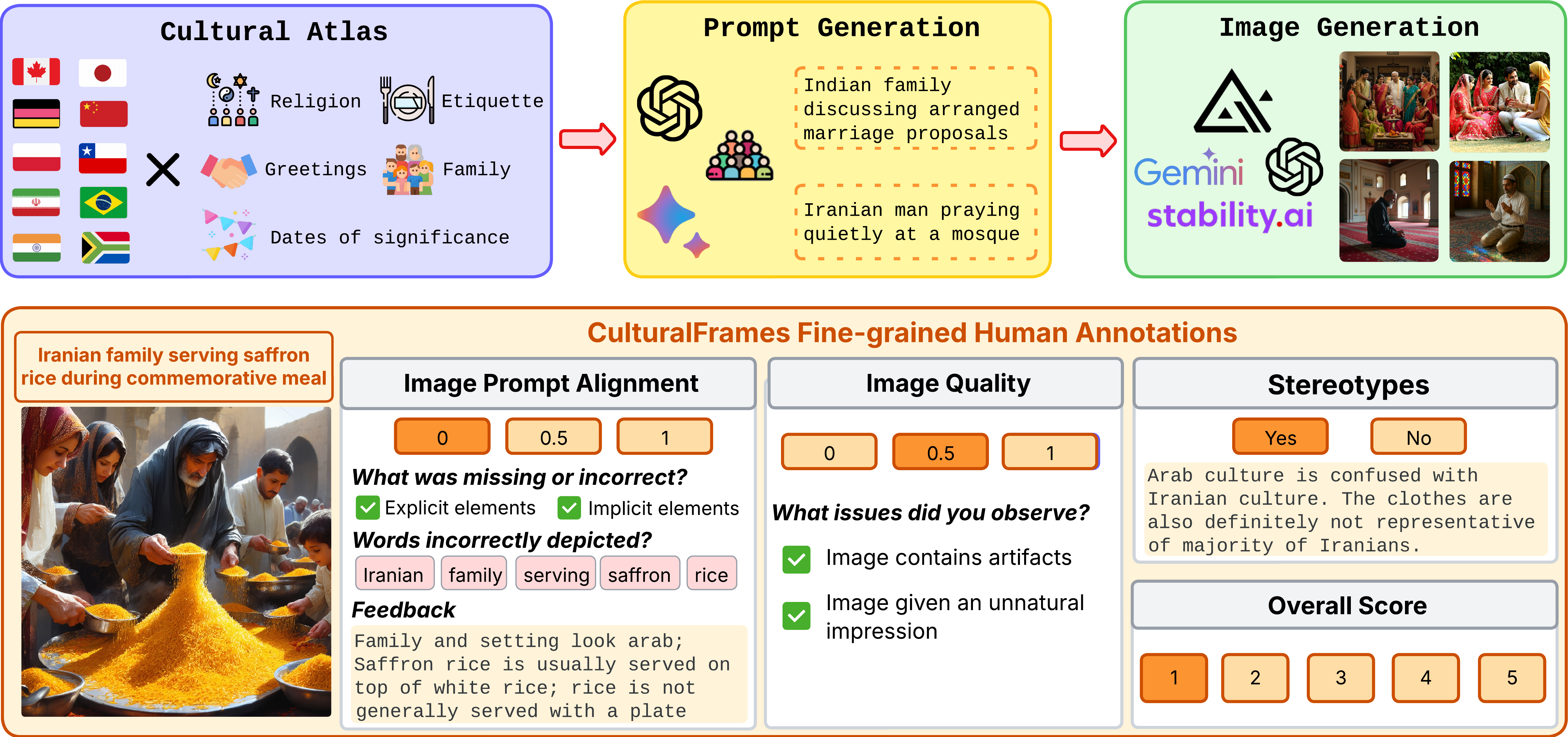}
    \caption{Overview of the \dataset\ dataset pipeline and annotation process. Prompts are first generated using cultural assertions from the Cultural Atlas across categories such as religion and family (top-left). These are transformed into culturally grounded textual prompts using large language models and human filtering (top-middle), and then rendered into images using state-of-the-art text-to-image models (top-right). Human annotators provide fine-grained evaluations (bottom) across four axes: image-prompt alignment, image quality, stereotype presence, and overall score, along with detailed feedback highlighting cultural inaccuracies and visual artifacts.}
    \label{fig:pipeline}
\end{figure*}

\section{\dataset}

We detail our data collection pipeline below and highlight the design decisions that make it distinct from standard annotation efforts. 

\subsection{Selection of Countries} 
We operationalize cultural groups using countries as a proxy~\citep{adilazuarda2024measuringmodelingculturellms}, building upon the premise that individuals within a country share a substantial amount of common cultural knowledge, implicit understandings, and norms that shape their daily interactions and practices~\citep{hofstede2010cultures,hershcovich-etal-2022-challenges}.
To create a dataset with diverse cultures, we selected countries spanning five continents and representing diverse cultural zones as per the zone categorization in the World Values Survey (WVS; \citealt{haerpfer2022world}).
Thus, our selection includes countries from the following cultural zones: West and South Asia (India), Confucian (China, Japan), African-Islamic regions (Iran, South Africa), Latin America (Brazil, Chile), English-speaking (Canada), Catholic Europe (Poland), and Protestant Europe (Germany).\footnote{We acknowledge that the labels assigned to these cultural categories are limited in their precision. Yet, these categories present the cross-cultural variation relevant to this work.}

\subsection{Selection of Cultural Categories} 
\label{sec:cultural_categories}
Our dataset is designed to evaluate culturally relevant expectations in visual generations. Specifically, we target five socio-cultural domains from CulturalAtlas~\citep{mosaica2024culturalatlas} deeply embedded in day-to-day life: 1) family, addressing familial roles, hierarchy, and interactions; 2) greetings, covering norms in social and business interactions; 3) etiquette, involving conduct during visits, meals, gift-giving, etc.; 4) religion, reflecting rituals and customs shaping group identities; 5) and dates of significance, highlighting celebrations of cultural, historical, or religious importance.
These categories were selected due to their coverage in the CulturalAtlas for the selected countries and their potential to induce prompts that elicit both explicit (\textit{i.e.}, elements directly mentioned in the prompt) and implicit (\textit{i.e.}, not mentioned in the prompt but inferred from shared cultural commonsense and needed for cultural authenticity) cultural expectations.\looseness=-1

\subsection{Data Generation Pipeline}

Building on the cultural categories, we first generate culturally grounded prompts reflecting the core values described above. 
For each prompt, we generate corresponding images and evaluate across multiple dimensions from culturally knowledgeable annotators to assess whether T2I models capture both explicit and implicit cultural expectations. \Cref{fig:pipeline} summarizes the data generation and human image annotation pipeline.

\paragraph{Prompt Generation.} We use Cultural Atlas~\citep{mosaica2024culturalatlas} as our knowledge base to extract cultural expectations (norms, practices, values) written as assertions. Cultural Atlas is an educational resource informed by extensive community interviews and validated by cultural experts. To generate culturally grounded prompts, we first extract concise assertions from Cultural Atlas content and feed them to GPT-4o~\citep{openai2024gpt4ocard} using designed instructions (see \Cref{app:prompt_generation}). These instructions guide the model to embed cultural expectations into the prompts for realistic and observable everyday scenarios. Next, we use GPT-4o~\citep{openai2024gpt4ocard} and Gemini~\citep{geminiteam2024gemini15unlockingmultimodal} to automatically validate the generated prompts, discarding any that are overly abstract, culturally misaligned, or not visually depictable. As a final step, we present each prompt to three culturally knowledgeable annotators. Only prompts agreed upon by the majority are retained in the dataset (more details in \Cref{app:prompt_filtering}). Example assertions and prompts from our benchmark are shown in \Cref{tab:example_prompts}. 

\begin{table}[!t]
    \centering
    \renewcommand{\arraystretch}{1.1} %
    \setlength{\tabcolsep}{4pt}
    \small 
    \begin{tabularx}{\columnwidth}{>{\raggedright\arraybackslash}p{4cm} 
                                     >{\raggedright\arraybackslash}X}
        \toprule
        \textbf{Assertion (CulturalAtlas)} & \textbf{Generated Prompts} \\
        \midrule 
        \texttt{Greetings (India)}: Indians expect people to greet the eldest or most senior person first. When greeting elders, some may touch the ground or the elder's feet as a sign of respect. & 
        \textbf{(1)} Grandchildren touching grandfather's feet at an Indian temple. \textbf{(2)} Indian village elder blessing children during harvest festival. \\
        \midrule
        \texttt{Religion (Iran)}: Most Iranians believe in Islam, but due to politicization, many younger citizens have withdrawn. Devout followers often practice privately at home. & 
        \textbf{(1)} Iranian family praying together at home. \textbf{(2)} Elderly Iranian man praying in a quiet mosque. \\
        \bottomrule
    \end{tabularx}
    \caption{Examples of assertions in CulturalAtlas for two categories \textit{greetings} in India and \textit{religion} in Iran and corresponding generated prompts.}
    \label{tab:example_prompts}
\end{table}

\paragraph{Image Generation.} We generate images using four state-of-the-art T2I models: two open-source models (Flux 1.0-dev~\citep{flux2024} and Stable Diffusion 3.5 Large (SD)~\citep{esser2024scalingrectifiedflowtransformers}) and two closed-source models (Imagen3~\citep{imagenteamgoogle2024imagen3} and GPT-Image~\citep{openai2025gpt4oimage}). We note that Imagen3 includes a prompt expansion mechanism, active by default.
To keep the evaluation practical and consistent across models, we generate one image per model per prompt. While this may appear limiting, our analysis (Appendix~\ref{app:single_image}) shows that output diversity across generations is generally low, and key issues identified by annotators tend to generalize across multiple outputs. In \Cref{fig:cultural_frames_examples}, we present prompt-image examples.

\paragraph{Rating Collection.}
We developed a human rating collection interface and the associated annotation guidelines. We tested several interface designs and variants of annotation guidelines to collect high-quality annotations. The final interface and the guidelines are provided in \Cref{app:image_rating}. To ensure high data quality, we filtered for attentive annotators and ensured a minimum of 25 unique, culturally knowledgeable workers\footnote{Annotators were selected based on the following criteria: born in the country, national of the country, have spent the majority of the first 18 years of life there, and are a resident of the country. The residency criterion was relaxed for China to ensure a sufficient annotator pool size.} per country. 
We collect data from three annotators for each country using the Prolific\footnote{\url{https://www.prolific.com/}} platform. 
Our annotation process captures detailed, multifaceted feedback. 
Each annotator first evaluates how well the image aligns with the prompt (\textbf{image-prompt alignment}), considering both explicit elements stated in the prompt and implicit elements expected based on cultural context. Following \citet{ku2024imagenhubstandardizingevaluationconditional}, we use a 3-point Likert scale: 0.0 (no alignment), 0.5 (partial), and 1.0 (complete). For scores below 1, annotators specify whether explicit, implicit, or both types of elements were missing or not depicted satisfactorily in the image, and highlight the specific words in the prompt whose visual depictions were not satisfactory, along with providing justifications for why they were not satisfactory. This fine-grained rating scheme allows us to analyze the interplay between various quality aspects and their relation with perceived cultural appropriateness. 
Annotators flag \textbf{stereotypes} in the images, providing justifications if present. Next, they assess \textbf{image quality}, noting issues such as distortions, artifacts, or unrealistic object rendering. Finally, they assign an \textbf{overall image score} on a 5-point Likert scale. See \Cref{fig:pipeline} (bottom) for an example of human annotation for different criteria for an image-prompt pair.\looseness=-1
\section{Data Analysis}

\paragraph{Prompts.}
\dataset\ consists of \nprompts\ prompts collected from 10 countries, with each country contributing between 90 and 110 prompts, ensuring balanced cross-country representation. The prompts are distributed across five cultural categories introduced in~\Cref{sec:cultural_categories}: etiquette (24.3\%), religion (14.4\%), family (14.2\%), greetings (13.1\%), and dates of significance (34\%). For a detailed per-country breakdown, see \Cref{fig:plots_grid} in \Cref{app:prompt_distribution}.

\paragraph{Images.}
We generate images for our prompt set using both open- and closed-source models. While open-source models produce an image for every prompt, the safety filters of closed-source models block a subset of generations. This issue is most noticeable with Imagen3, which filters out 290 prompts—29.5\% of the prompts, primarily due to policies against depicting children~\footnote{We requested an exemption from the provider to bypass these filters and will incorporate the missing images if access is granted}. For comparison, GPT-4o blocks only 5 prompts. In total, we collect 3,637 images.

\paragraph{Inter-rater Agreement.} 
We collect a total of \nannotations\ ratings, with each image rated by three annotators. To measure agreement among raters, we compute Krippendorff's alpha \citep{krippendorff2013content}, obtaining 0.32 for prompt alignment, 0.28 for image quality, and 0.36 for the overall score. These scores are comparable to, or better than, those reported in prior works evaluating cultural understanding in T2I models~\citep{kannen2025aestheticsculturalcompetencetexttoimage, bayramli2025diffusionmodelsgloballens}. A detailed comparison with prior works, along with potential factors influencing the agreement scores, is provided in Appendix~\ref{app:human_agreement}.

\paragraph{What aspect of the generated image dominates annotators' overall assessment?}
We find that the overall score given by annotators is strongly correlated with image–prompt alignment (Spearman rank correlation of 0.68), whereas image quality shows a more moderate correlation of 0.45. This trend holds consistently across countries, suggesting that annotators prioritize faithfulness to the prompt over aesthetic appeal when rating images. Also, stereotype is negatively correlated with overall score weakly (-0.21), which indicates a lower impact of the presence of stereotypes on overall score. Interestingly, the results contrast with findings from prior work using side-by-side image comparisons~\citep{kirstain2023pickapicopendatasetuser}, where image quality often dominates overall preference judgments. 
\begin{figure}[!t]
    \includegraphics[width=0.5\textwidth]{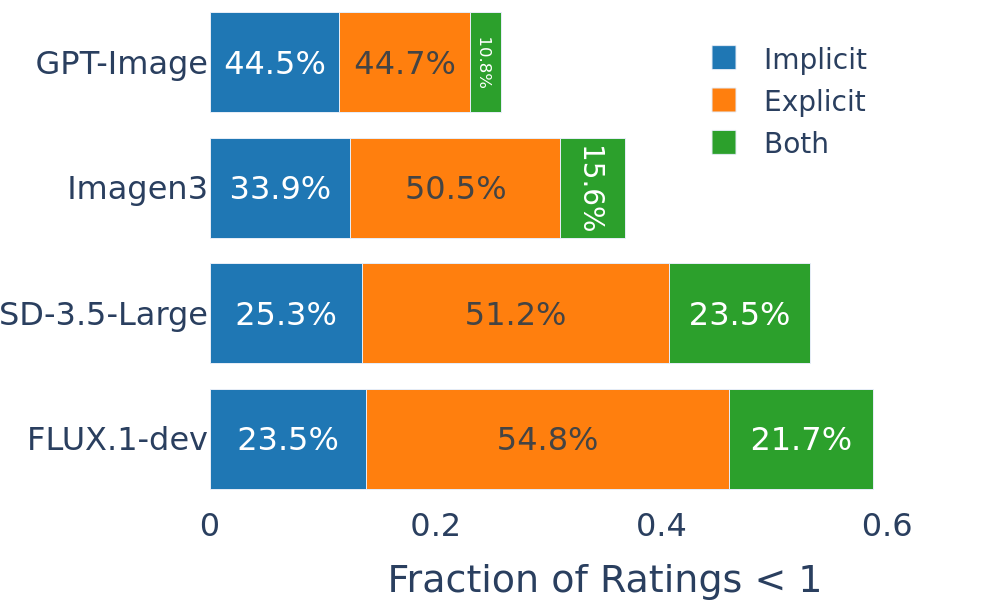}
    \caption{Distribution of image-prompt alignment errors (score <1) by model, grouped by error type: implicit, explicit, or both. Bar lengths show fraction of total errors; \% show each type's share of the model's total errors.}
    \label{fig:implicit_explicit}
\end{figure}

\begin{figure*}[t]
    \includegraphics[width=\textwidth]{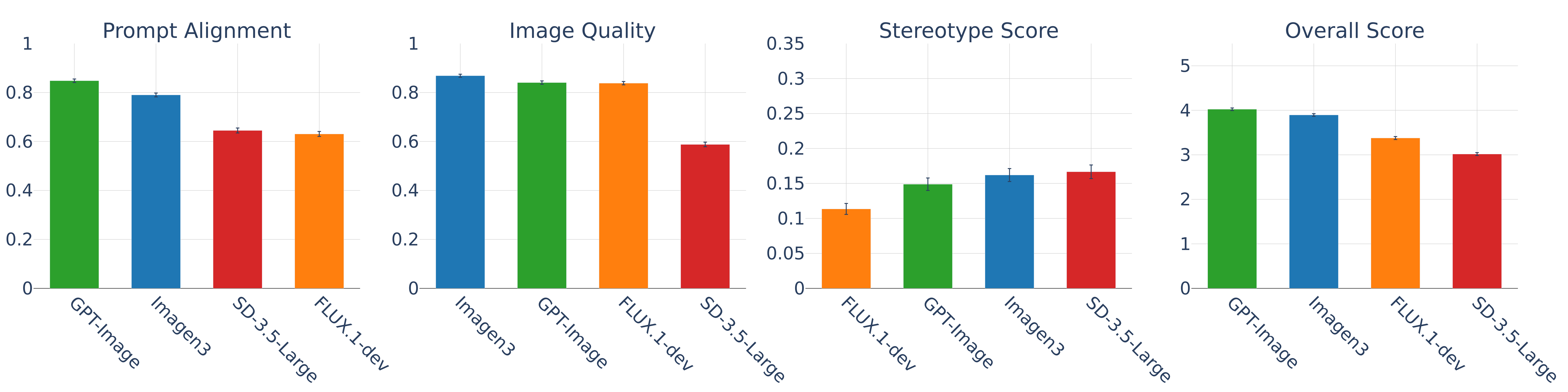}
    \caption{Human evaluation results for selected T2I models. From left to right: 1) Prompt Alignment ($0-1$ scale, $1=$perfect alignment). 2) Image Quality ($0-1$ scale, $1=$highest quality). 3) Stereotype Score ($0-1$ scale, $0$ indicates no stereotyping). 4) Overall Score ($1-5$ Likert scale, $5=$best overall). For fairness, we compare across prompts that have images generated by all models.}
    \label{fig:model_performance}
\end{figure*}

\begin{figure*}[t]
    \includegraphics[width=\textwidth]{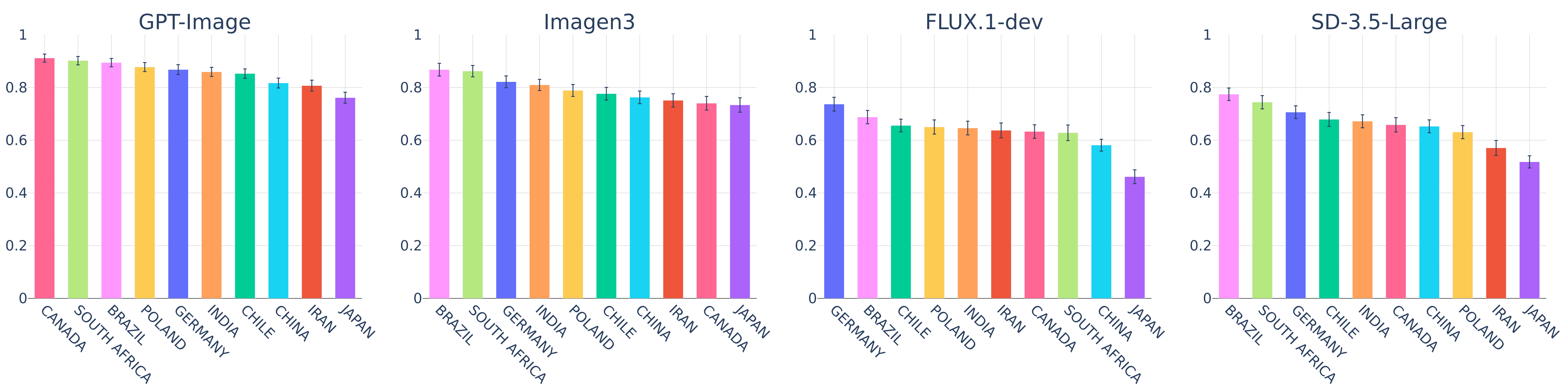}
    \caption{Prompt alignment scores across countries for a given model.}
    \label{fig:across_countries}
\end{figure*}

\section{Evaluating T2I Models on \dataset}

\paragraph{How do different models perform for different criteria across different countries?}

\Cref{fig:model_performance} shows human evaluation results for prompt alignment, image quality, stereotype, and overall score. We find that GPT-Image achieves the highest prompt alignment ($0.85$), followed by Imagen3 ($0.79$). The open-source models, SD-3.5-Large and Flux, fall behind with scores of $0.66$ and $0.63$, respectively. For image quality, Imagen3 is rated highest, with GPT-Image and Flux performing comparably well. 
SD-3.5-Large, however, scores far behind the other models. 
Across all models, including the state-of-the-art closed-source ones, the proportion of images rated stereotypical ranged from 10\% to 16\%, with SD-3.5-Large generating stereotypical visuals the most and Flux the least. Overall, raters prefer images from GPT-Image, consistent with the prompt alignment result. SD received the lowest overall score, most likely due to poorer image quality and higher stereotype levels, despite outperforming Flux in prompt alignment.
Our findings (\Cref{fig:across_countries} and \Cref{fig:all_criteria_model_comparision}) indicate notable cross-country variations in both the overall score and perceived importance of different evaluation criteria. For instance, even assessments of image quality differ, showing a discernible trend where Asian countries tend to assign lower scores across multiple criteria.\looseness=-1

\paragraph{Is there a preferred model across countries?} 
For prompt alignment (see \Cref{fig:model_order}), GPT-Image is consistently preferred across countries, followed by Imagen3. Among open-source models, SD-3.5-Large is generally more faithful except for Germany, Poland, and Iran, where Flux performs better. 
In \Cref{fig:all_criteria_model_comparision}, we show detailed results across countries and all categories.
Regarding image quality, Imagen3 is the preferred model, likely due to its hyper-realistic generations. Interestingly, concerning stereotypes, closed-source models are ranked as more stereotypical for 6 out of the 10 countries.  

\paragraph{Which aspect---implicit or explicit---do models fail to capture, and is this consistent across countries?}
Across \dataset, annotators gave sub-perfect scores (below 1) for 44\% of the time. Out of these, 50.3\% are attributed to issues with explicit elements, 31.2\% to implicit elements, and 17.9\% to both. While explicit errors are most common, implicit cultural failures still account for 49.1\% of these cases, underscoring persistent challenges in capturing culturally nuanced, context-dependent knowledge. 
\Cref{fig:implicit_explicit} shows that GPT-Image has the lowest overall image-prompt alignment error rate (ratings < 1), with its errors roughly evenly split between implicit and explicit types. In contrast, other models, particularly SD-3.5-Large and FLUX, exhibit higher total error rates where explicit errors form the largest share of their respective alignment failures.
These results indicate that improvements are needed in both explicit and implicit cultural modeling.\looseness=-1

In Canada, Poland, Germany, and Brazil, approximately two‑thirds of comments mention explicit prompt mismatches, indicating that literal fidelity dominates their feedback. Conversely, annotator feedback from India, China, and South Africa is more evenly distributed, with roughly half of the remarks targeting explicit flaws and half targeting implicit cultural elements. At the opposite end of the spectrum, annotators from Japan and Iran predominantly highlight implicit cultural elements, such as absent rituals, attire, or local setting, with only about one‑third of their comments citing explicit tokens. Chile follows the latter trend, albeit less strongly. Collectively, these observations indicate that T2I models increasingly fail to capture users' implicit cultural expectations in regions like Asia and the Middle East, as contrasted with user feedback from the Americas and Europe.

\paragraph{Which words do models most frequently misinterpret?}
\Cref{fig:wordcloud} displays every word in the prompt that at least one rater labeled as erroneous, revealing two striking patterns. First, country demonyms (e.g., Iranian, Brazilian, Chinese, Japanese) are prominent. A closer examination of the rater comments reveals these words are typically highlighted as errors for two reasons: (i) a country‑specific element is missing from the image, or (ii) the annotators are not able to relate to the depicted content. Second, terms such as \textit{family}, \textit{festival}, \textit{ceremony}, \textit{wedding}, \textit{temple}, \textit{meal}, \textit{guests}, \textit{tea}, \textit{greeting}, \textit{music}, \textit{costumes}, and \textit{flags} account for much of the remaining error frequency. These words represent broad cultural signifiers---rituals, social roles, and iconic objects---indicating that T2I models frequently misrepresent such elements.

\paragraph{What do annotators flag in low‑quality images?}
When images received low quality scores, annotators most often selected the presence of artifacts 70.4\% of the time and the image having an unnatural impression 50.9\% of the time on average. Across models, SD-3.5-Large accounts for the largest share of both artifact flags (54.4\%) and ``unnatural'' flags (43.2\%). Notably, Flux-1.0-dev and GPT-Image also show high ``unnatural'' shares ($\approx$ 24\% and $\approx$ 22\%, respectively). Our qualitative analysis indicates that ``unnatural'' is typically triggered by global coherence issues where scenes or cultural elements seem implausible for the cultural setting, whereas ``artifacts'' reflects local distortions (e.g., blur, distortions).

\begin{figure}[!t]
    \centering
    \includegraphics[width=0.47\textwidth]{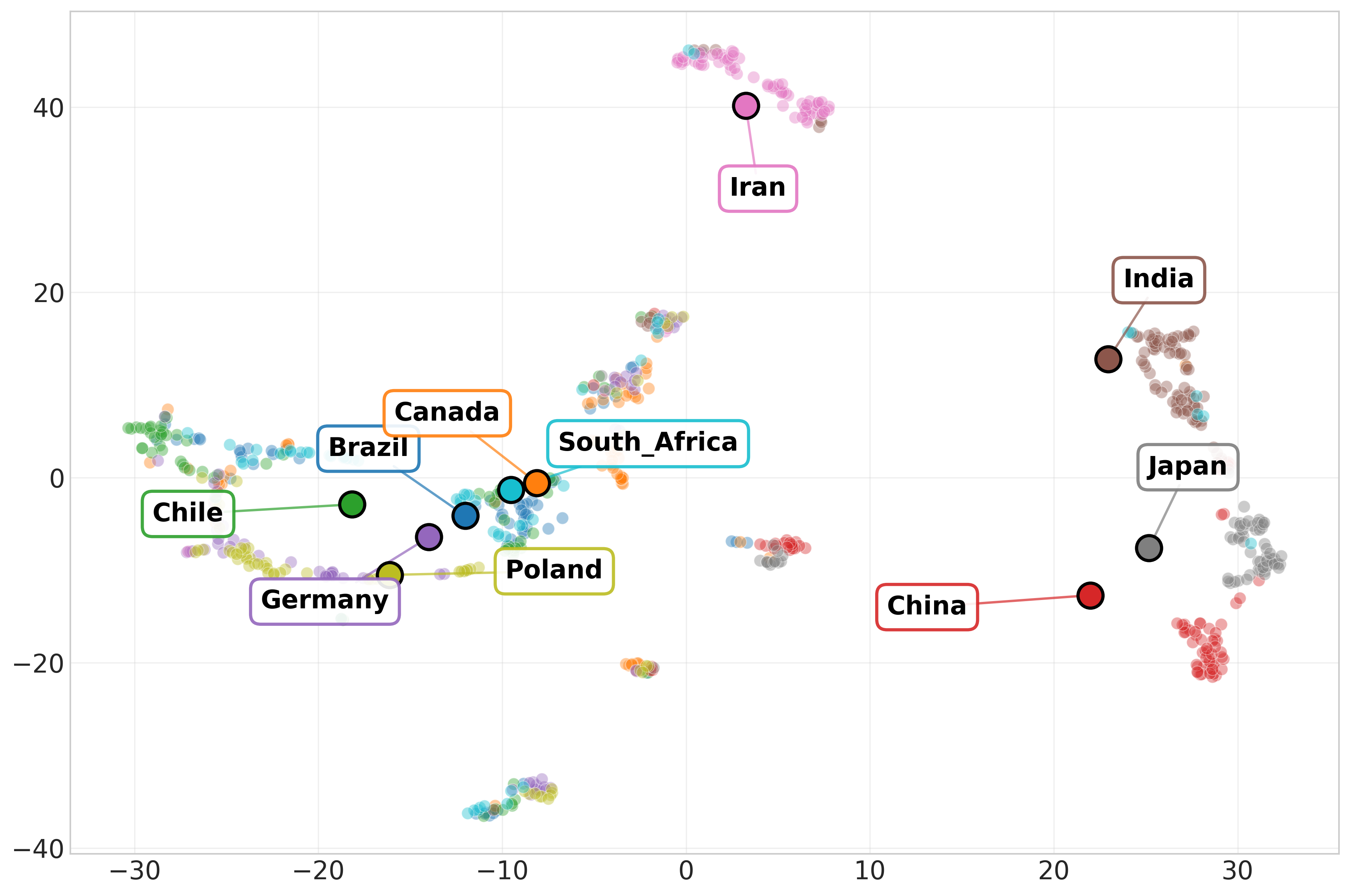}
    \caption{tSNE plot of Imagen3 images. Labeled markers show image embedding centroids per country.}
    \label{fig:tsne}
\end{figure}

\begin{figure*}[!h]
    \centering
    \includegraphics[width=1\textwidth]{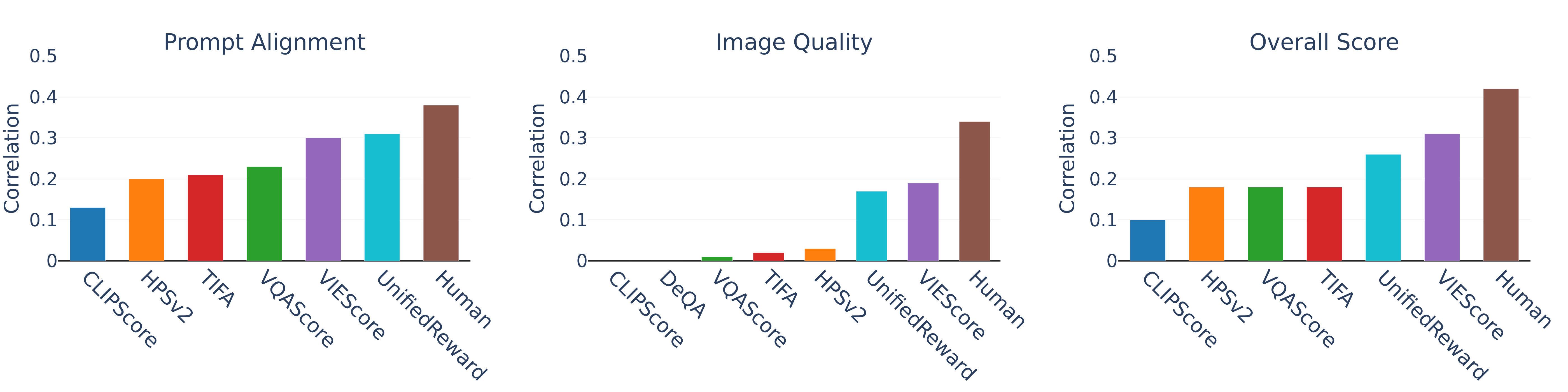}
    \caption{Spearman rank correlation of various T2I evaluation metrics with human ratings across three criteria: prompt alignment, image quality, and overall score. Human denotes the human-human Spearman rank correlation.}
    \label{fig:metric_corelations}
\end{figure*}

\paragraph{In what way do models fail across different countries?}
To identify reasons behind model failures, we analyze free-form comments collected from annotators. For each country, we embed the comments using a sentence transformer\footnote{\url{https://huggingface.co/sentence-transformers/all-mpnet-base-v2}} and cluster them using HDBScan~\cite{hdbscan}. We then prompt GPT-4o to summarize each cluster with a concise label and explanations. This approach reveals distinct failure patterns across regions. In Asia, models frequently misrepresent traditions and religious practices, often relying on stereotypes. In African contexts, outputs lacked cultural authenticity, defaulting to generic or Westernized portrayals. South American outputs suffered from poor regional specificity and inaccurate depictions of people's appearances. 
Similarly, Canadian content lacked appropriate demographic diversity and Indigenous representation. Further, we investigate the nature of the generated images by embedding them using the CLIP vision encoder.\footnote{\url{https://huggingface.co/openai/clip-vit-large-patch14}} As shown in \Cref{fig:tsne}, images generated by Imagen3 for Asian countries form distinct clusters, while those from other regions lack such clear grouping. This finding is corroborated by annotators in Europe and South America, who struggle to identify country-specific visual cues in generated images, indicating that the model fails to capture cultural distinctiveness.\looseness=-1

\section{Evaluating T2I Metrics on \dataset}

\paragraph{Metrics analyzed.}  
We analyze six representative metrics, each reflecting a different evaluation paradigm:
CLIPScore \citep{hessel-etal-2021-clipscore} is an embedding-based metric that computes cosine similarity between CLIP embeddings of the image and prompt.
HPSv2 \citep{wu2023humanpreferencescorev2} enhances CLIPScore by fine-tuning the CLIP model on human preference data.
TIFA \citep{hu2023tifaaccurateinterpretabletexttoimage} uses a VQA-based framework to assess faithfulness. We use GPT-4o-mini for question generation and Qwen2.5-VL-32B-Instruct~\citep{qwen2.5-VL} as the answering model.
VQAScore \citep{vqascore}, UnifiedReward \citep{wang2025unifiedrewardmodelmultimodal}, and VIEScore \citep{viescore} leverage vision-language models to evaluate generated images. For VQAScore, we leverage the CLIP-FlanT5 model introduced in the original VQAScore paper, use UnifiedReward-qwen-7B based on Qwen2.5-VL-7B for UnifiedReward, and use GPT-4o as VLM for VIEScore, which provides both a score and a textual reason for its assessment. Finally, we evaluate DeQA \citep{deqa_score}, a VLM trained specifically for image‑quality assessment.

\paragraph{How do metrics perform against different rating criteria?}  
We evaluate how well current T2I metrics correlate with human judgments across prompt alignment, image quality, and overall score (see \Cref{fig:metric_corelations}). UnifiedReward, an open‑source reward model, slightly edges the best closed‑model setup, VIEScore, on prompt alignment, achieving a Spearman correlation of 0.31 compared to 0.30 for the latter. While this is below the human-human agreement of 0.38, it notably outperforms all other metrics. In contrast, TIFA exhibits a lower correlation, potentially because it only accounts for explicit elements mentioned in the prompt. This highlights a gap between metric design and alignment with human perception.
The performance gap is even more pronounced for \emph{image quality}, where all metrics correlate poorly with human ratings. Nevertheless, VIEScore again performs best, followed closely by UnifiedReward. The relatively stronger performance of HPSv2 may be attributed to its training on image pairs, with human preference likely driven by image quality, potentially making it more sensitive to visual appeal. By contrast, DeQA, despite being trained specifically for image‑quality assessment on standard IQA datasets, shows near‑zero correlation ($\approx 0.0$) on our benchmark, likely due to domain and distribution shift between the data used to train DeQA and \dataset. Taken together, the overall weak correlations suggest that current metrics fail to capture the subjective nature of image quality as assessed by humans.
For the \emph{overall score}, VIEScore again demonstrates the highest alignment with human judgments, achieving a correlation of 0.31 (human–human: 0.42), with UnifiedReward close behind. Notably, HPSv2, despite being trained on human preferences, shows relatively poor performance, likely due to limited annotator and prompt diversity in the human preference dataset it was trained on. CLIPScore consistently underperforms, indicating limitations as a general-purpose evaluation metric, particularly for culturally sensitive image assessments. Overall, these results suggest that VLM‑based metrics, such as VIEScore and UnifiedReward, have the upper hand in capturing culturally grounded human.\looseness=-1

\paragraph{Do explanations provided by VLM-based metrics capture the mistakes human raters highlight?}
To further analyze the effectiveness of the overall best-performing metric on our benchmark, VIEScore, we evaluate whether its generated explanations reflect the issues raised by human annotators. We consider only cases where at least two annotators flagged mistakes with substantiated reasons. We adopt an LLM‑as‑a‑judge setup, instructing it to assess the alignment between VIEScore's reasoning and human concerns on a 1–5 Likert scale. The instructions are shown in \Cref{fig:llm_judge}. To mitigate potential model biases, we collect scores from 4 different LLMs -- GPT-4o \citep{openai2024gpt4ocard}, Gemini 2.5 Flash \citep{geminiteam2024gemini15unlockingmultimodal}, Claude3.5-Sonnet \citep{anthropic2024claude35addendum}, DeepSeek-Chat \citep{deepseekai2025deepseekv3technicalreport} -- and aggregate them per instance.
To calibrate the LLM's judgments, we provided five in-context examples corresponding to varying quality levels. Additionally, we manually review 100 judge-provided scores, sampled across countries, confirming that the judges produce consistent, high‑quality assessments.
The results reveal that VIEScore's explanations achieve an average rating of 2.19/5 (std: 1.19), indicating only partial overlap with human rationale. These findings suggest that current metrics have substantial room to improve alignment with human judgments and reasoning. Some qualitative examples are provided in \Cref{tab:human-metric-rationale}.\looseness=-1
\section{Discussion}

Based on our analysis of cultural misalignment in text-to-image models and their evaluation metrics, we highlight three key directions for improvement.

\paragraph{Can culturally informed prompt expansion improve cultural alignment?} CulturalFrames prompts are concise, leaving many cultural aspects implicit for the model to infer. For example, the prompt "a bride and groom exchanging vows at their Hindu wedding" omits scene elements like the priest or the presence of the sacred fire, which are essential for faithful depiction. To examine whether making such cues explicit can improve generations, we build on our analysis of model failures and develop a prompt-expansion method that addresses recurrent omissions such as cultural objects, family members/roles, setting details, and mood/atmosphere. We select the 20 lowest‑scoring prompts per country (200 total across 10 countries) and expand each prompt with Gemini‑2.5‑Flash \citep{comanici2025gemini25pushingfrontier} (see \Cref{app:prompt_expansion} for instructions). We then generate images with Flux.1-Dev, the strongest open‑source model in our study, and evaluate image–prompt alignment with VIEScore~\citep{viescore}, the metric that best correlates with human judgments. Prompt expansion improves the overall VIEScore from 7.3 to 8.4, showing that targeted, culturally informed expansion helps models attend to cues humans care about. More broadly, this highlights how \dataset\ and our fine-grained analysis can guide the design of prompt-expansion methods.\looseness=-1

\paragraph{Can we improve metric performance through explicit instructions?}

Current T2I metrics are not explicitly guided to consider implicit and explicit prompt elements when evaluating image alignment. To test whether such guidance improves performance, we modify the instructions given to GPT-4o within VIEScore, replacing them with novel annotation guidelines we developed for human raters (see \Cref{fig:new_viescore}). We then re-evaluate images for image-prompt alignment using this instruction-augmented version of the VIEScore.
This intervention yields measurable gains in correlation with human ratings, with the Spearman correlation increasing from 0.30 to 0.32. We conduct a bootstrap significance test and confirm the improvement is significant at 95\% confidence. We also see an improvement in alignment of explanations with human rationales under the same LLM‑as‑a‑judge setup, increasing from 2.19 to 2.37 on a 5‑point scale.
These results show that careful, culturally informed instruction design can move the needle on both scores and rationales, indicating that part of the gap stems from missing guidance rather than model capacity alone. Nonetheless, the metric's reasoning still falls considerably short of human rationale, pointing to the need for richer cultural knowledge and training beyond prompt design.\looseness=-1

\paragraph{Does explicit training of VLMs to judge images improve culturally aligned evaluation?}
Current VLMs used for evaluation are typically not explicitly trained to judge images, raising the question of whether such training could improve cultural alignment. To investigate this, we compare UnifiedReward~\citep{wang2025unifiedrewardmodelmultimodal}, built on Qwen2.5-VL-7B Instruct and trained on diverse human-annotated multimodal preference and scoring data, with its backbone model. While the UnifiedReward training covers varied content, it does not specifically target cultural scenarios. Across all criteria, UnifiedReward shows markedly higher correlations with human judgments: image–prompt alignment (0.31 vs. 0.17), image quality (0.17 vs. 0.01), and overall score (0.28 vs. 0.14). Notably, it even surpasses GPT-4o-based VIEScore in image–prompt alignment (0.31 vs. 0.30). These results indicate that preference-based judge training, despite being agnostic to cultural content, can meaningfully enhance the cultural alignment of metric scores, aligning with prior evidence that such training benefits VLM-based evaluators~\citep{li2025vlrewardbenchchallengingbenchmarkvisionlanguage}.\looseness=-1
\section{Conclusions}

In this work, we introduce \dataset, a novel benchmark comprising \nprompts\ cultural prompts, \nimages\ generated images, and \nannotations\ human annotations, spanning ten countries and five socio-cultural domains. \dataset\ assesses the ability of T2I models to generate images across diverse cultural contexts. We find that state-of-the-art T2I models not only fail to meet the more nuanced implicit expectations, but also the less challenging explicit expectations. In fact, models fail to meet cultural expectations 44\% of the time on average across countries. Failures to meet explicit expectations averaged a surprisingly high 68\% across models and countries, with implicit expectation failures also significant at 49\%.
Finally, we demonstrate that existing T2I evaluation metrics correlate poorly with human judgments of cultural alignment.\looseness=-1

\section{Limitations}
Our study faces limitations due to our data collection methods and the scope of the \dataset. We approximated cultural groups as countries for annotator recruitment, which may potentially oversimplify cultural identities and conflate culture with nationality due to practical constraints like information available in CulturalAtlas and annotator availability.

Our strategic choice to maximize diversity by recruiting multiple annotators per country, while enriching the evaluation with varied viewpoints, inherently presents a trade-off. A broader range of interpretations, stemming from a more diverse group, can naturally lead to lower inter-rater agreement scores when compared to evaluations conducted by a smaller, more homogenous annotator pool.
It is this trade-off, coupled with the inherent subjectivity of the task, that provides context for our inter-annotator agreement results. This reflects the inherent subjectivity of evaluating cultural nuances and expectations. 

A further limitation, driven by practical considerations of scale, is a generation of only a single image per model per prompt. This single-instance evaluation makes it challenging for annotators to definitively identify stereotypical associations, as patterns of representation across multiple generations for the same prompt cannot be observed.

\section{Ethical Considerations}

Our \dataset\ benchmark comprises prompts and generated images, whose cultural alignment is rated by professional annotators via Prolific from the relevant countries.
To ensure wide cultural representation, we recruited annotators from three distinct community groups within these countries, compensating them at \$10-15 per hour for all tasks performed, a rate established after pilot testing. This reflects our commitment to fair and inclusive data collection practices.

Despite the efforts, we acknowledge a key limitation: equating cultural groups with national borders within or across these national lines. This simplification may overlook the complex realities of minority and diaspora communities. We thus urge future research to explore finer-grained distinctions within cultural groups. While recognizing these constraints, we are hopeful that our work contributes to a deeper understanding of cultural nuances in visual generations and provides a foundation for such future investigations.

\section{Acknowledgements}

We would like to thank Saba Ahmadi, Qian Yang, Ankur Sikarwar and Rohan Banerjee for their help with early pilots for prompt generation and image rating. We also thank the Mila IDT team for their technical support and for managing the computational resources. Additionally, Aishwarya Agrawal received support from the Canada CIFAR AI Chair award throughout this project. 
Karolina
Sta\'nczak was supported by the Mila P2v5 grant, the Mila-Samsung grant, and by an ETH AI Center postdoctoral fellowship.
This project was generously funded by a research grant from Google. This project was also supported by funding from IVADO and the Canada First Research Excellence Fund.

\bibliography{custom}

\newpage

\appendix
\onecolumn{

\noindent\textbf{\Large Appendix}\\

\vspace{8pt} 

\noindent\textbf{\Large Table of Contents}

\begin{flushright}
    \textbf{Page}
\end{flushright}

\noindent
\renewcommand{\arraystretch}{1.2}
\begin{tabularx}{\linewidth}{Xr} 
    \textbf{A. \dataset} \dotfill & \pageref{app:culturalframes} \\  
    \hspace{2em} A.1 Prompt Generation \dotfill & \pageref{app:prompt_generation} \\  
    \hspace{2em} A.2 Prompt Filtering \dotfill & \pageref{app:prompt_filtering} \\
    \hspace{2em} A.3 Prompt Distribution Across Categories \dotfill & \pageref{app:prompt_distribution} \\
    \hspace{2em} A.4 Image Generation \dotfill & \pageref{app:image_generation} \\
    \hspace{2em} A.5 Prompt-Image Examples \dotfill & \pageref{app:prompt_image_examples} \\
    \hspace{2em} A.6 Single Image Generation Analysis \dotfill & \pageref{app:single_image} \\
    \hspace{2em} A.7 Inter Human Agreement \dotfill & \pageref{app:human_agreement} \\[0.5em]

    \textbf{B. Image Rating} \dotfill & 
    \pageref{app:image_rating} \\ 
    \hspace{2em} B.1 Rating Interface \dotfill & \pageref{app:image_rating} \\ 
    \hspace{2em} B.2 Annotator Demnographics \dotfill & \pageref{app:annotator_demographics} \\[0.5em]

    \textbf{C. Text-to-Image Models' Analysis} \dotfill & \pageref{fig:model_order} \\
    \hspace{2em} C.1 Prompt Expansion Case Study \dotfill & \pageref{fig:prompt_expansion}\\
    \hspace{2em} C.2 Model Ranking by Countries and Criteria \dotfill & \pageref{fig:model_order} \\
    \hspace{2em} C.3 Model Scores by Country and Criteria  \dotfill & \pageref{fig:country_analysis_overall_score}\\
    \hspace{2em} C.4 Word Cloud of Annotator-Flagged Issues by Country  \dotfill & \pageref{fig:wordcloud}\\[0.5em]

    \textbf{D. Text-to-Image Metrics' Analysis} \dotfill & \pageref{fig:llm_judge} \\ 
    \hspace{2em} D.1 LLM-as-a-Judge Evaluation Protocol \dotfill & \pageref{fig:llm_judge} \\
    \hspace{2em} D.2 Qualitative Examples of VIEScore Failures \dotfill & \pageref{tab:human-metric-rationale} \\
    \hspace{2em} D.3 Revised Evaluation Instructions Given to VIEScore \dotfill & \pageref{fig:new_viescore} \\
\end{tabularx}
}
\newpage

\begin{table*}[t]
\centering
\begin{tabular}{lrrrrr}
\toprule
\textbf{Country} & \textbf{Unique Annotators} & \textbf{Avg Age} & \textbf{\% Male} & \textbf{\% Female} & \textbf{\% Other} \\
\midrule
Brazil        & 35  & 36.1 & 69.0 & 31.0 & 0.0 \\
Canada        & 34  & 37.9 & 47.9 & 52.1 & 0.0 \\
Chile         & 35  & 31.1 & 77.7 & 22.3 & 0.0 \\
China         & 40  & 33.0 & 32.3 & 67.7 & 0.0 \\
Germany       & 51  & 35.1 & 68.5 & 31.5 & 0.0 \\
India         & 32  & 31.7 & 46.6 & 53.4 & 0.0 \\
Iran          & 28  & 32.0 & 47.0 & 53.0 & 0.0 \\
Japan         & 25  & 44.2 & 56.1 & 40.6 & 3.2 \\
Poland        & 27  & 32.0 & 62.0 & 38.0 & 0.0 \\
South Africa  & 83  & 32.9 & 35.1 & 64.9 & 0.0 \\
\bottomrule
\end{tabular}
\caption{Summary of participant demographics by country.}
\label{tab:demographics_summary}
\end{table*}

\section{\dataset}\label{app:culturalframes}

This section outlines the full pipeline used to create the \dataset. We describe how culturally grounded prompts were generated, filtered, and verified by human annotators across multiple countries. We also detail how these prompts were used to generate images from various text-to-image models, along with the settings and parameters used for generation.

\subsection{Prompt Generation} \label{app:prompt_generation}

We begin with the Cultural Atlas~\cite{mosaica2024culturalatlas}, a curated knowledge base of cross-cultural attitudes, practices, norms, behaviors, and communication styles, designed to inform and educate the public about Australia's migrant populations. The Atlas provides detailed textual descriptions across categories such as family structures, greeting customs, cultural etiquette, religious beliefs, and more. 
We use the Cultural Atlas as a source of culturally grounded information to guide prompt generation. However, not all categories in the Atlas are suitable for visual depiction. We selected five categories—\textit{dates-of-significance}, \textit{etiquette}, \textit{family}, \textit{religion}, and \textit{greetings}—based on two main criteria: (1) the content describes values or practices that can be meaningfully represented in images, and (2) the category is consistently available across a broad set of countries to support cross-cultural comparison.

We parsed the textual content from each selected category and segmented it into paragraphs using newline characters. Each paragraph served as an input “excerpt” to an LLM for prompt generation.
Given a country and an excerpt, we prompted GPT-4o (gpt-4o-2024-08-06)~\citep{openai2024gpt4ocard} to generate two short prompts (each under 15 words) that: (i) were grounded in the excerpt's content, (ii) described a culturally relevant and visually observable scenario, and (iii) included sufficient country-specific context, either explicitly or implicitly. The prompts were designed to reflect underlying cultural values through everyday, observable situations, such as a wedding ceremony or a workplace interaction. To guide this process, we crafted category-specific instructions that encouraged the model to generate meaningful and culturally grounded prompts.

We began by generating a small number of prompts per category, which were evaluated by human annotators to assess whether the scenarios were both visually depictable and culturally appropriate (see Section~\ref{app:prompt_filtering} for details). Prompts that passed these quality checks were reused as few-shot in-context examples to guide further prompt generation. This iterative process enabled us to scale prompt creation while maintaining cultural fidelity and diversity. Instructions provided to GPT-4o~\citep{openai2024gpt4ocard} used across different categories are provided in Figures \ref{fig:greeting_prompt}, \ref{fig:religion_prompt}, \ref{fig:etiquette_prompt}, \ref{fig:family_prompt}, \ref{fig:dates_prompt}.

\begin{figure*}[ht]
  \includegraphics[width=\textwidth]{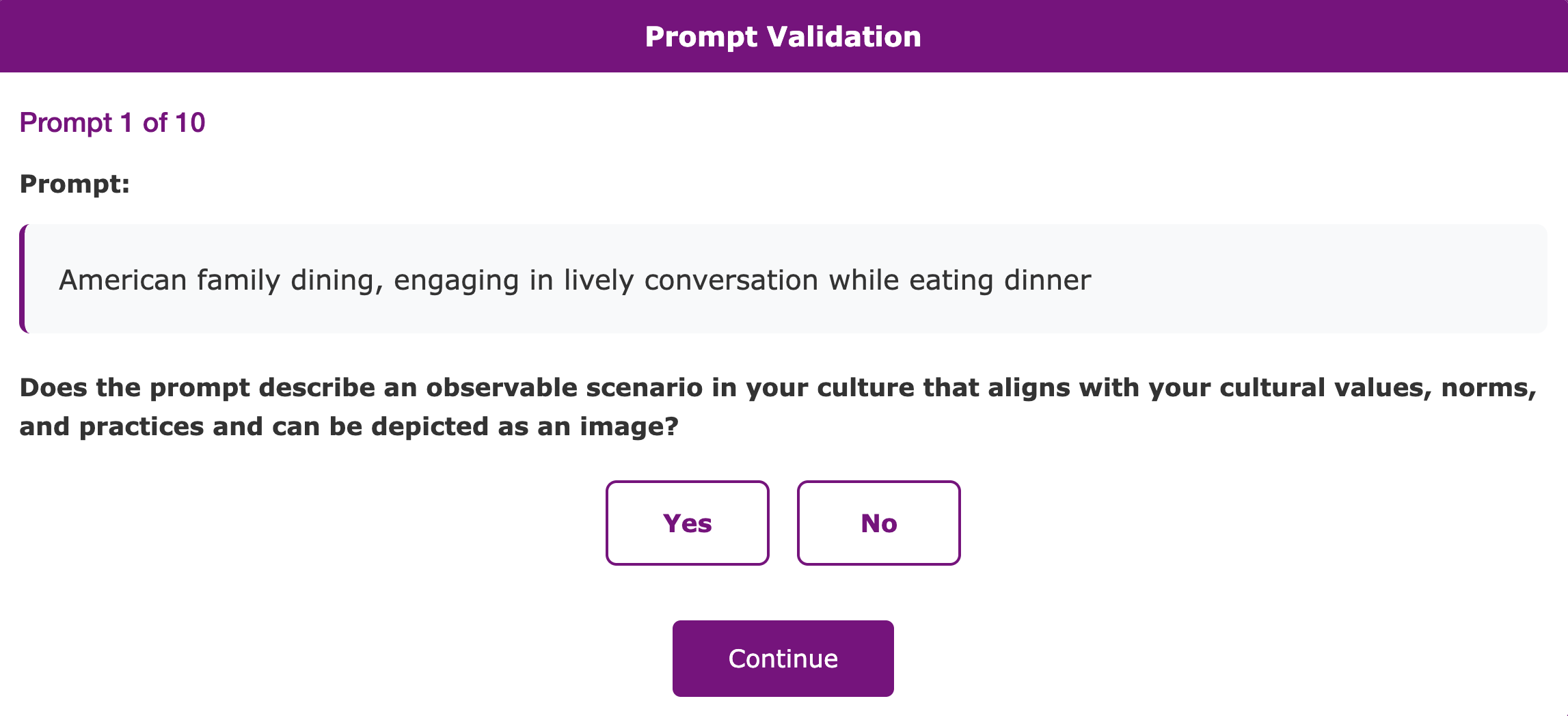}
    \caption{Prompt filtering interface where annotators choose ``Yes/No'' for a given prompt depending on whether the prompt reflects an observable scenario in their culture that aligns with their cultural values.}
    \label{fig:inst_prompt_filtering}
\end{figure*}

\vspace{0.5em}
\begin{figure*}[t]
\centering
\small 
\begin{tcolorbox}[title=Prompt Instructions (Greeting), myboxstyle]

\textbf{Purpose:}

We want to test whether text-to-image models can accurately capture a country's distinct greeting practices. You will be given:

\begin{enumerate}
    \item A country name  
    \item A short excerpt on greeting norms: an implicit description of how people in this country typically greet each other, or some information that relates to greeting customs.
\end{enumerate}

\textbf{Your Task:}

Use these inputs to produce two short prompts (each under 15 words) that is rooted in the provided excerpt and explore diverse scenarios, to evaluate the image-generation model's understanding of the greeting values and norms. Each prompt should:

\begin{itemize}
    \item Be clearly rooted in the excerpt's details and context (e.g., setting, participants, timing). You must not deviate from the provided excerpt.  
    \item Represent a social scenario or interaction where the greeting norm or value mentioned in the excerpt can be observed. These should be concrete, observable situations that commonly occur in this culture/country.  
    \item Be diverse, realistic scenario, and under 15 words  
    \item Be visually depictable - that is, it must be possible to generate a meaningful and culturally relevant image based on the prompt. This includes avoiding verbal greetings that cannot be depicted in the image.  
\end{itemize}

\textbf{Important:}  
Make sure the country can be inferred from the prompt. It should be either stated explicitly like mentioning a region or name of the country or there must be enough country specific elements in the prompt to infer the country.\\

\textbf{Note:}  
If the information provided cannot be used to create a practical observable scenario that can be depicted as an image, return "N/A".\\

\textbf{Return the prompts in this JSON format:}
\begin{verbatim}
{
    "prompt_1": "...",
    "prompt_2": "..."
}
\end{verbatim}

\textbf{Here are the inputs:}
\begin{itemize}
    \item Country: \texttt{\{country\}}  
    \item Excerpt: \texttt{\{excerpt\}}
\end{itemize}

\textbf{Previously Generated Prompts (to avoid duplication):}
\begin{verbatim}
{already_generated_prompts}
\end{verbatim}

\textbf{Accepted Examples:}
\begin{verbatim}
{incontext_examples_positive}
\end{verbatim}

\textbf{Rejected Examples:}
\begin{verbatim}
{incontext_examples_negative}
\end{verbatim}

Generate \textbf{exactly two} new prompts that satisfy all of the criteria above, follow the style/patterns of the accepted examples, avoid the issues shown in the rejected ones, and explore diverse scenarios different from the ones already generated. Output \textbf{only} the JSON object specified.

\end{tcolorbox}
\caption{Instructions used to generate prompts for the greeting category}
\label{fig:greeting_prompt}
\end{figure*}

\begin{figure*}[t]
\centering
\small 
\begin{tcolorbox}[title=Prompt Instructions (Religion), myboxstyle]

\textbf{Purpose:}

We want to test whether text-to-image models can accurately capture how religion is practiced in a particular country along with its norms, practices, rituals, traditions, and values. You will be given:

\begin{enumerate}
    \item A country name  
    \item A short excerpt on religious norms: an implicit description of how religion is practiced or influences everyday life, or some information that is related to religious practices.
\end{enumerate}

\textbf{Your Task:}

Use these inputs to produce two short prompts (each under 15 words) that is rooted in the provided excerpt and explore diverse scenarios, to evaluate the image-generation model's understanding of the religion of the country. Each prompt should:

\begin{itemize}
    \item Be clearly rooted in the excerpt's details and context (e.g., setting, participants, timing). You must not deviate from the provided excerpt  
    \item Create prompts that describe specific daily interactions, rituals, or scenarios that reflect the cultural values and social norms related to religion and mentioned in the excerpt. These should be concrete, observable situations that commonly occur in this culture/country.  
    \item Be diverse, realistic scenario, and under 15 words  
    \item Be visually depictable - that is, it must be possible to generate a meaningful and culturally relevant image based on the prompt.  
\end{itemize}

\textbf{Important:}  
Make sure the country can be inferred from the prompt. It should be either stated explicitly like mentioning a region or name of the country or there must be enough country specific elements in the prompt to infer the country.\\

\textbf{Note:}  
If the information provided cannot be used to create a practical observable scenario that can be depicted as an image, return "N/A".\\

\textbf{Return the prompts in this JSON format:}
\begin{verbatim}
{
    "prompt_1": "...",
    "prompt_2": "..."
}
\end{verbatim}

\textbf{Here are the inputs:}
\begin{itemize}
    \item Country: \texttt{\{country\}}  
    \item Excerpt: \texttt{\{excerpt\}}
\end{itemize}

\textbf{Previously Generated Prompts (to avoid duplication):}
\begin{verbatim}
{already_generated_prompts}
\end{verbatim}

\textbf{Accepted Examples:}
\begin{verbatim}
{incontext_examples_positive}
\end{verbatim}

\textbf{Rejected Examples:}
\begin{verbatim}
{incontext_examples_negative}
\end{verbatim}

Generate \textbf{exactly two} new prompts that satisfy all of the criteria above, follow the style/patterns of the accepted examples, avoid the issues shown in the rejected ones, and explore diverse scenarios different from the ones already generated. Output \textbf{only} the JSON object specified.
\end{tcolorbox}
\caption{Instructions used to generate prompts for the religion category}
\label{fig:religion_prompt}
\end{figure*}

\begin{figure*}
\centering
\small
\begin{tcolorbox}[title=Prompt Instructions (Etiquette), myboxstyle]

\textbf{Purpose:}

We want to test whether text-to-image models can accurately capture how etiquette is practiced in a particular country, including norms, manners, and social conduct related to visiting, gifting, eating, and other social situations. You will be given:

\begin{enumerate}
    \item A country name  
    \item A short excerpt on etiquette norms: an implicit description of how people in this country engage with each other in different social situations, or some information related to etiquette.
\end{enumerate}

\textbf{Your Task:}

Use these inputs to produce two short prompts (each under 15 words) that is rooted in the provided excerpt and explore diverse scenarios, to evaluate the image-generation model's understanding of etiquette. Each prompt should:

\begin{itemize}
    \item Be clearly rooted in the excerpt's details and context (e.g., setting, participants, timing). You must not deviate from the provided excerpt  
    \item Represent a social scenario or interaction where the etiquette norm or value mentioned in the excerpt can be observed. It must be a realistic, observable scenario that commonly occurs in this culture/country.  
    \item Do not explicitly name the etiquette rule. Be implicit in conveying the details. The goal is to create situations where the etiquette rule can be observed and inferred by the model.
    \item Be diverse, realistic scenario, and under 15 words
    \item Be visually depictable - that is, it must be possible to generate a meaningful and culturally relevant image based on the prompt.
    \item Avoid using phrases like "arrving late", "arriving on time" and other such phrases that cannot be visualized in the image.
\end{itemize}

\textbf{Important:}  
Make sure the country can be inferred from the prompt. It should be either stated explicitly like mentioning a region or name of the country or there must be enough country specific elements in the prompt to infer the country.\\

\textbf{Note:}  
If the information provided cannot be used to create a practical observable scenario that can be depicted as an image, return "N/A".\\

\textbf{Return the prompts in this JSON format:}
\begin{verbatim}
{
    "prompt_1": "...",
    "prompt_2": "..."
}
\end{verbatim}

\textbf{Here are the inputs:}
\begin{itemize}
    \item Country: \texttt{\{country\}}  
    \item Excerpt: \texttt{\{excerpt\}}
\end{itemize}

\textbf{Previously Generated Prompts (to avoid duplication):}
\begin{verbatim}
{already_generated_prompts}
\end{verbatim}

\textbf{Accepted Examples:}
\begin{verbatim}
{incontext_examples_positive}
\end{verbatim}

\textbf{Rejected Examples:}
\begin{verbatim}
{incontext_examples_negative}
\end{verbatim}

Generate \textbf{exactly two} new prompts that satisfy all of the criteria above, follow the style/patterns of the accepted examples, avoid the issues shown in the rejected ones, and explore diverse scenarios different from the ones already generated. Output \textbf{only} the JSON object specified.
\end{tcolorbox}
\caption{Instructions used to generate prompts for the etiquette category}
\label{fig:etiquette_prompt}
\end{figure*}

\begin{figure*}
\centering
\small
\begin{tcolorbox}[title=Prompt Instructions (Family), myboxstyle]

\textbf{Purpose:}

We want to test whether text-to-image models can accurately depict how family values, structures, and dynamics operate in a particular country. You will be given:

\begin{enumerate}
    \item A country name  
    \item A short excerpt on family norms: an implicit description of how family life, roles, or relationships function in this culture.
\end{enumerate}

\textbf{Your Task:}

Use these inputs to produce two short prompts (each under 12 words) that are clearly rooted in the provided excerpt and explore diverse scenarios, to evaluate a model's understanding of these family practices. Each prompt should:

\begin{itemize}
    \item Be firmly based on the excerpt's context. You must not deviate from the provided excerpt  
    \item Portray family related interactions that happen in the culture/country conditioned on the values, norms provided in the excerpt  
    \item Avoid explicitly naming the core family norm or value, but include enough detail for the model to infer it  
    \item Depict diverse, realistic scenarios that convey familial interactions, each under 12 words
    \item Be visually depictable - that is, it must be possible to generate a meaningful and culturally relevant image based on the prompt.
\end{itemize}

\textbf{Important:}  
Make sure the country can be inferred from the prompt. It should be either stated explicitly like mentioning a region or name of the country or there must be enough country specific elements in the prompt to infer the country.\\

\textbf{Note:}  
If the information provided cannot be used to create a practical observable scenario that can be depicted as an image, return "N/A".\\

\textbf{Return the prompts in this JSON format:}
\begin{verbatim}
{
    "prompt_1": "...",
    "prompt_2": "..."
}
\end{verbatim}

\textbf{Here are the inputs:}
\begin{itemize}
    \item Country: \texttt{\{country\}}  
    \item Excerpt: \texttt{\{excerpt\}}
\end{itemize}

\textbf{Previously Generated Prompts (to avoid duplication):}
\begin{verbatim}
{already_generated_prompts}
\end{verbatim}

\textbf{Accepted Examples:}
\begin{verbatim}
{incontext_examples_positive}
\end{verbatim}

\textbf{Rejected Examples:}
\begin{verbatim}
{incontext_examples_negative}
\end{verbatim}

Generate \textbf{exactly two} new prompts that satisfy all of the criteria above, follow the style/patterns of the accepted examples, avoid the issues shown in the rejected ones, and explore diverse scenarios different from the ones already generated. Output \textbf{only} the JSON object specified.
\end{tcolorbox}
\caption{Instructions used to generate prompts for the family category}
\label{fig:family_prompt}
\end{figure*}

\begin{figure*}
\centering
\small
\begin{tcolorbox}[title=Prompt Instructions (Dates-of-significance), myboxstyle]

\textbf{Purpose:}

We want to test whether text-to-image models can accurately depict how a country observes its significant dates—festivals, holidays, or other notable events. You will be given:

\begin{enumerate}
    \item A country name  
    \item A short excerpt on a date of significance: an implicit description of festivities, traditions, or commemorative practices related to this important day.
\end{enumerate}

\textbf{Your Task:}

Use these inputs to produce two short prompts (under 12 words) that are clearly rooted in the provided excerpt and explore diverse scenarios, to evaluate a model's understanding of these celebrations. Each prompt should:

\begin{itemize}
    \item Be firmly based on the excerpt's context. You must not deviate from the provided excerpt  
    \item Represent daily interactions, rituals, or scenarios that are related to this date of significance. It must be a realistic, observable scenario that commonly occurs in this culture/country.  
    \item Convey the date of significance through rituals, traditions, or celebrations that are specific to this date.  
    \item Depict diverse, realistic scenarios that convey how people observe this date, each under 12 words.
    \item Be visually depictable - that is, it must be possible to generate a meaningful and culturally relevant image based on the prompt.
\end{itemize}

\textbf{Important:}  
Make sure the country can be inferred from the prompt. It should be either stated explicitly like mentioning a region or name of the country or there must be enough country specific elements in the prompt to infer the country.\\

\textbf{Note:}  
If the information provided cannot be used to create a practical observable scenario that can be depicted as an image, return "N/A".\\

\textbf{Return the prompts in this JSON format:}
\begin{verbatim}
{
    "prompt_1": "...",
    "prompt_2": "..."
}
\end{verbatim}

\textbf{Here are the inputs:}
\begin{itemize}
    \item Country: \texttt{\{country\}}  
    \item Excerpt: \texttt{\{excerpt\}}
\end{itemize}

\textbf{Previously Generated Prompts (to avoid duplication):}
\begin{verbatim}
{already_generated_prompts}
\end{verbatim}

\textbf{Accepted Examples:}
\begin{verbatim}
{incontext_examples_positive}
\end{verbatim}

\textbf{Rejected Examples:}
\begin{verbatim}
{incontext_examples_negative}
\end{verbatim}

Generate \textbf{exactly two} new prompts that satisfy all of the criteria above, follow the style/patterns of the accepted examples, avoid the issues shown in the rejected ones, and explore diverse scenarios different from the ones already generated. Output \textbf{only} the JSON object specified.
\end{tcolorbox}
\caption{Instructions used to generate prompts for the dates of significance category}
\label{fig:dates_prompt}
\end{figure*}

\subsection{Prompt Filtering} \label{app:prompt_filtering}
For every country, we ask 3 culturally knowledgeable annotators if the prompt represents a scenario observable in their culture and aligns with their values. Only those prompts that 2 or more annotators choose make it into \dataset. In \Cref{fig:inst_prompt_filtering}, we present the prompt filtering interface where annotators choose ``Yes/No'' for a given prompt depending on whether the prompt reflects an observable scenario in their culture that aligns with their cultural values.

\subsection{Prompt Distribution Across Categories} \label{app:prompt_distribution}

\begin{figure*}[!htb]
    \centering

    \begin{subfigure}[t]{0.3\textwidth}
        \centering
        \includegraphics[width=\linewidth]{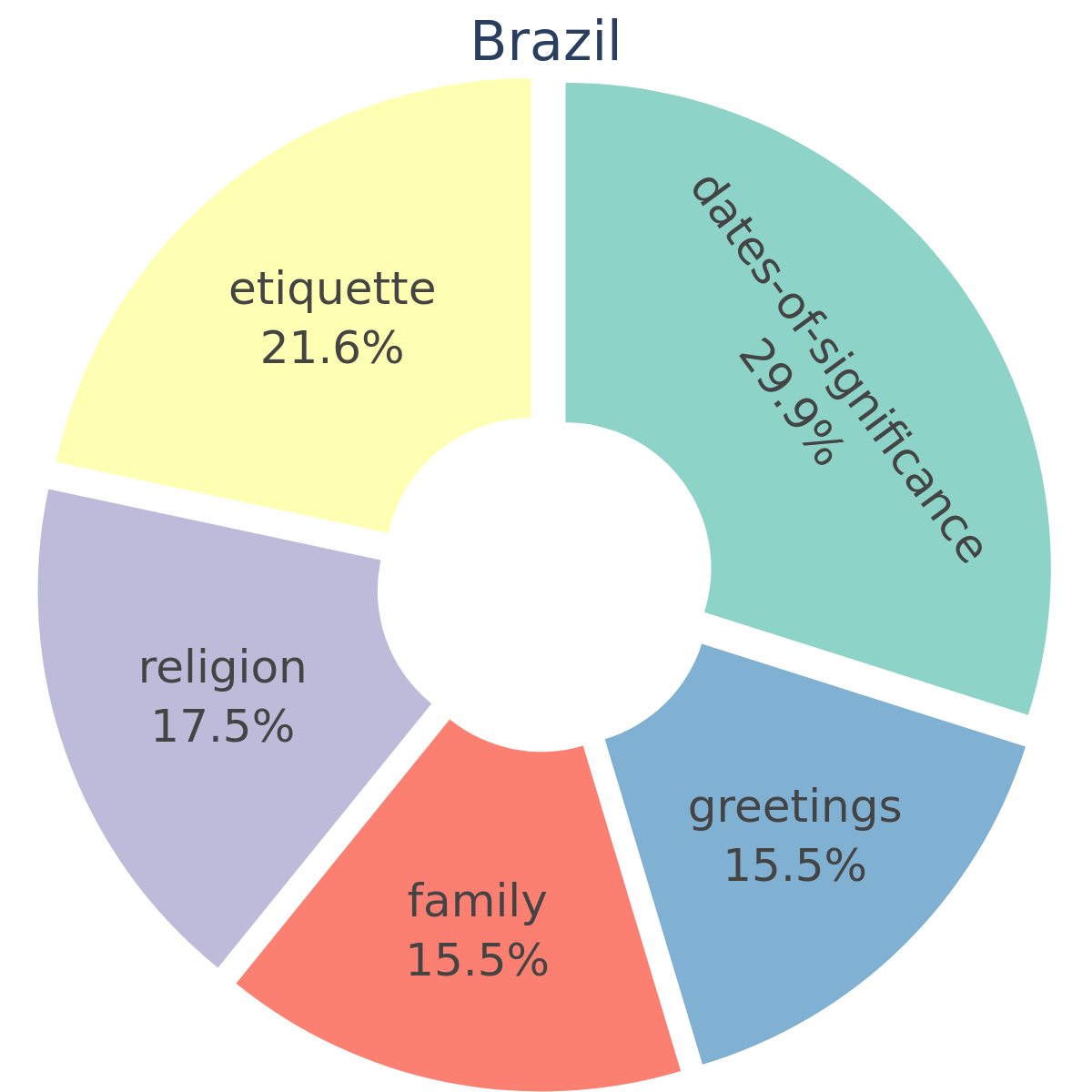}
    \end{subfigure}
    \hfill
    \begin{subfigure}[t]{0.3\textwidth}
        \centering
        \includegraphics[width=\linewidth]{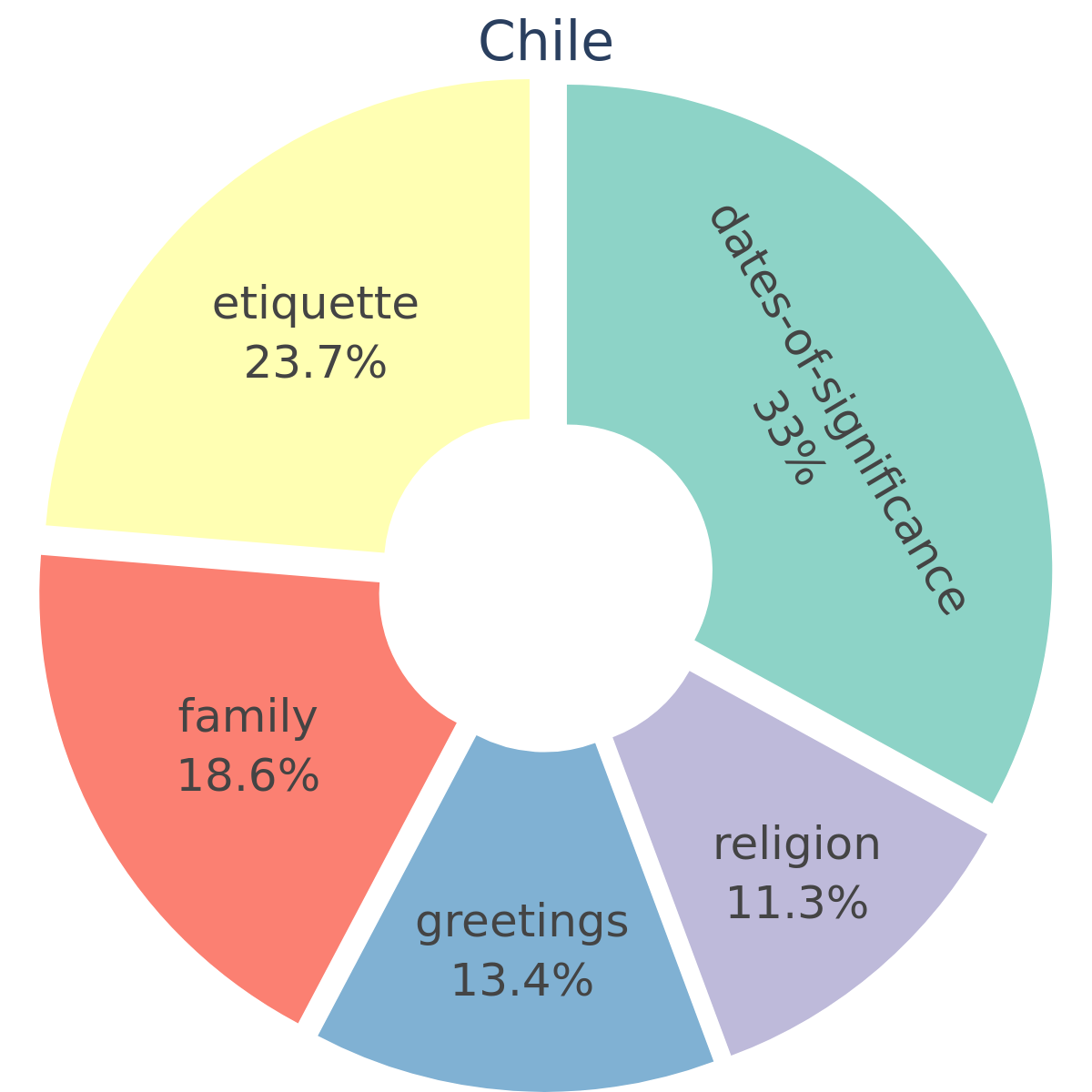}
    \end{subfigure}
    \hfill
    \begin{subfigure}[t]{0.3\textwidth}
        \centering
        \includegraphics[width=\linewidth]{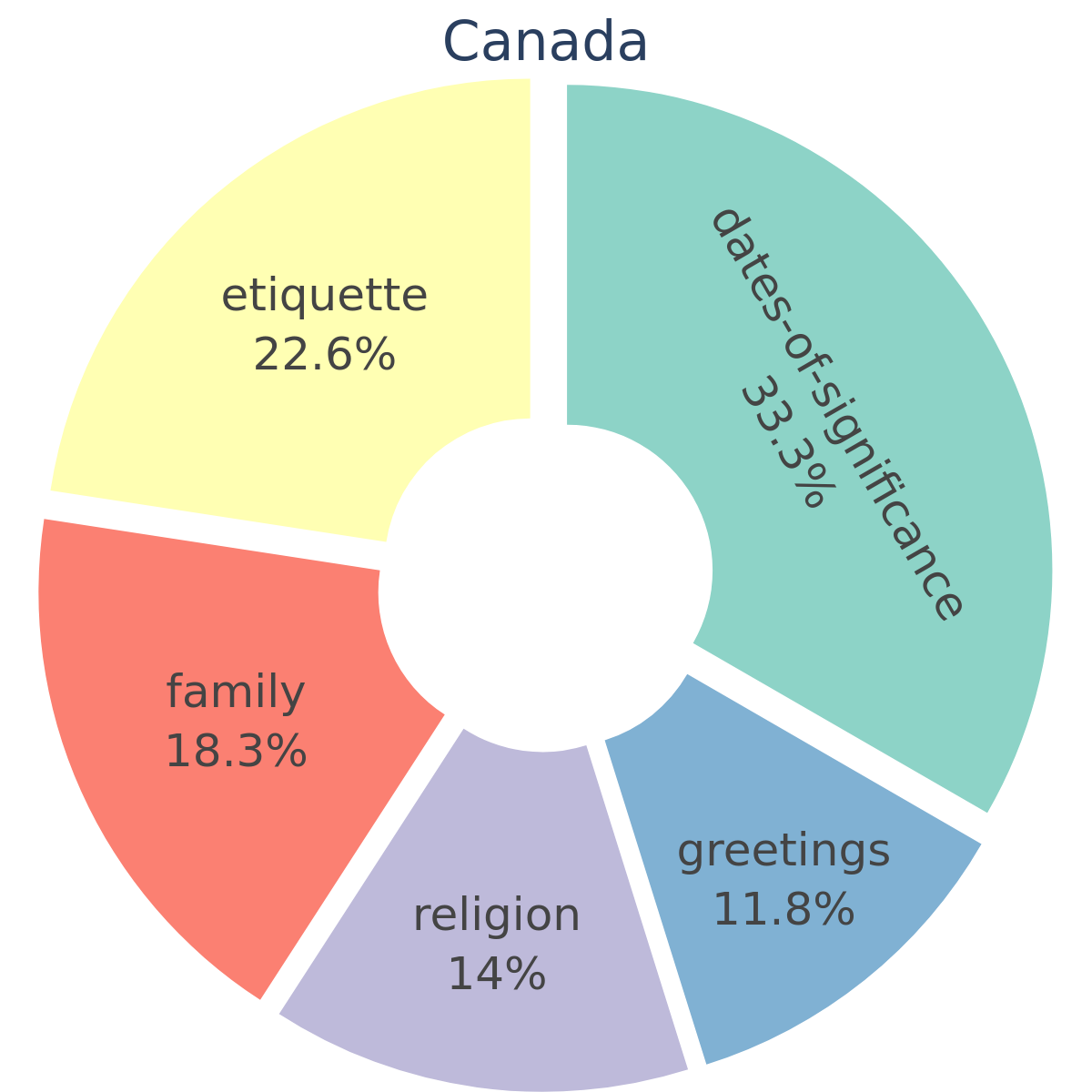}
    \end{subfigure}

    \vspace{1em}

    \begin{subfigure}[t]{0.3\textwidth}
        \centering
        \includegraphics[width=\linewidth]{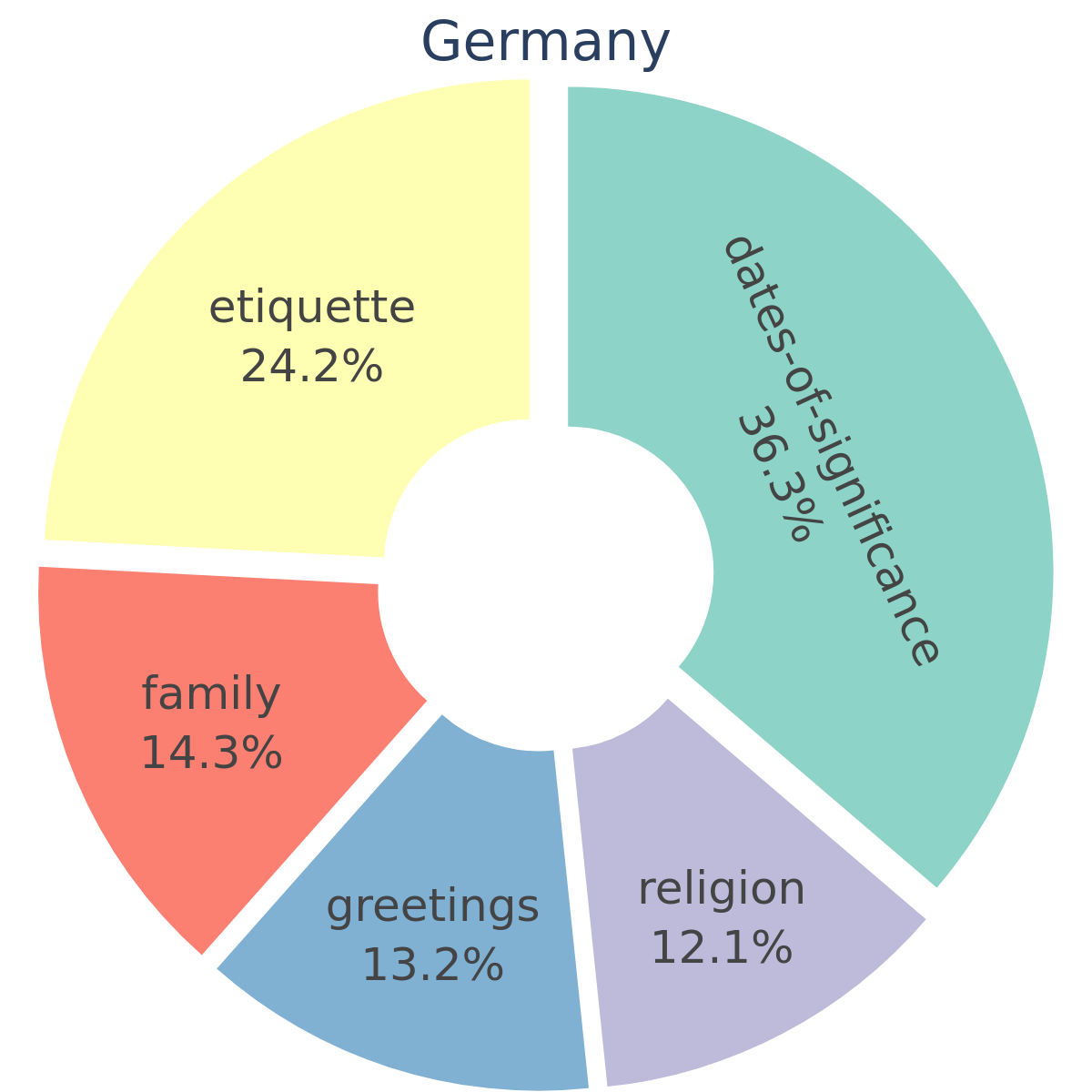}
    \end{subfigure}
    \hfill
    \begin{subfigure}[t]{0.3\textwidth}
        \centering
        \includegraphics[width=\linewidth]{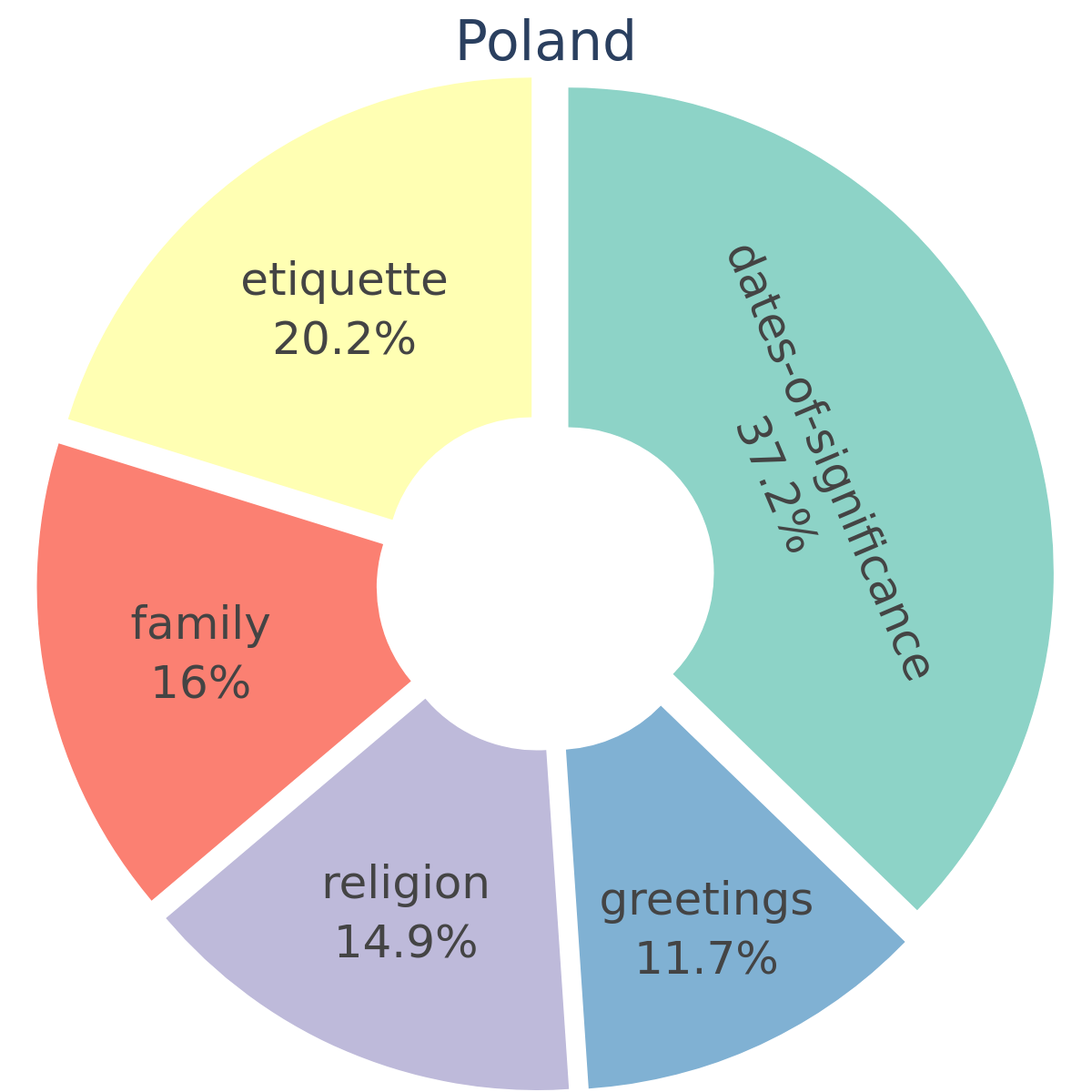}
    \end{subfigure}
    \hfill
    \begin{subfigure}[t]{0.3\textwidth}
        \centering
        \includegraphics[width=\linewidth]{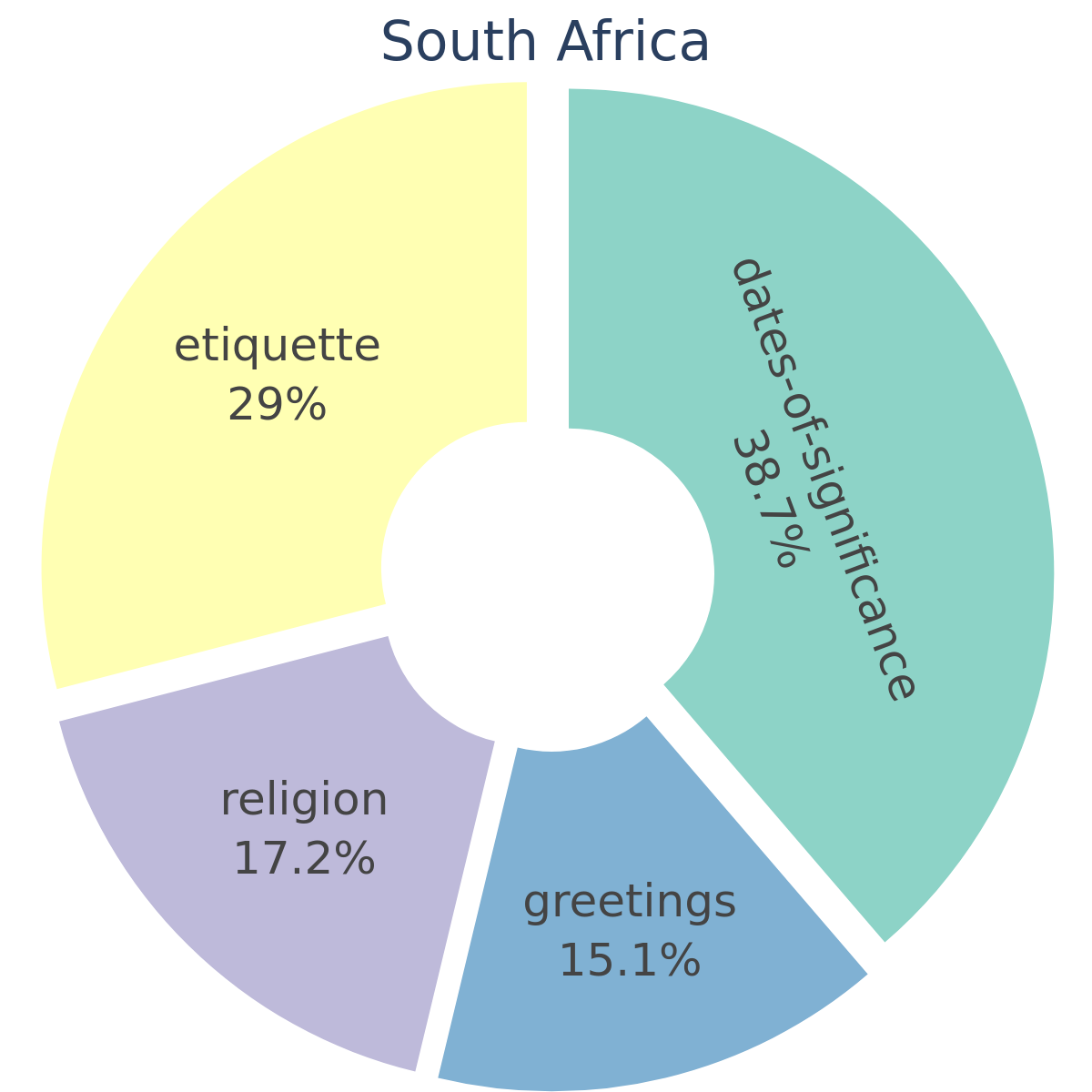}
    \end{subfigure}

    \vspace{1em}

    \begin{subfigure}[t]{0.3\textwidth}
        \centering
        \includegraphics[width=\linewidth]{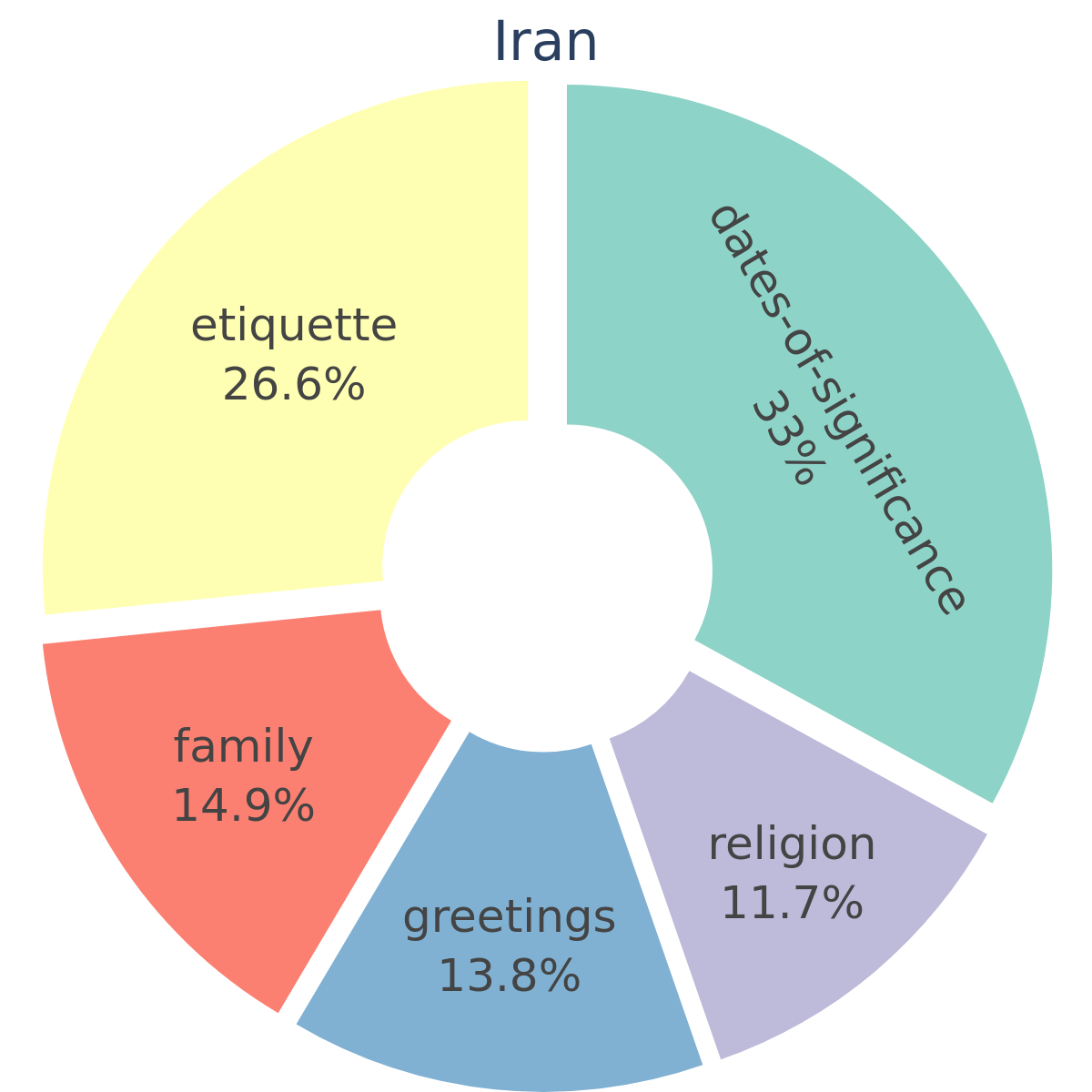}
    \end{subfigure}
    \hfill
    \begin{subfigure}[t]{0.3\textwidth}
        \centering
        \includegraphics[width=\linewidth]{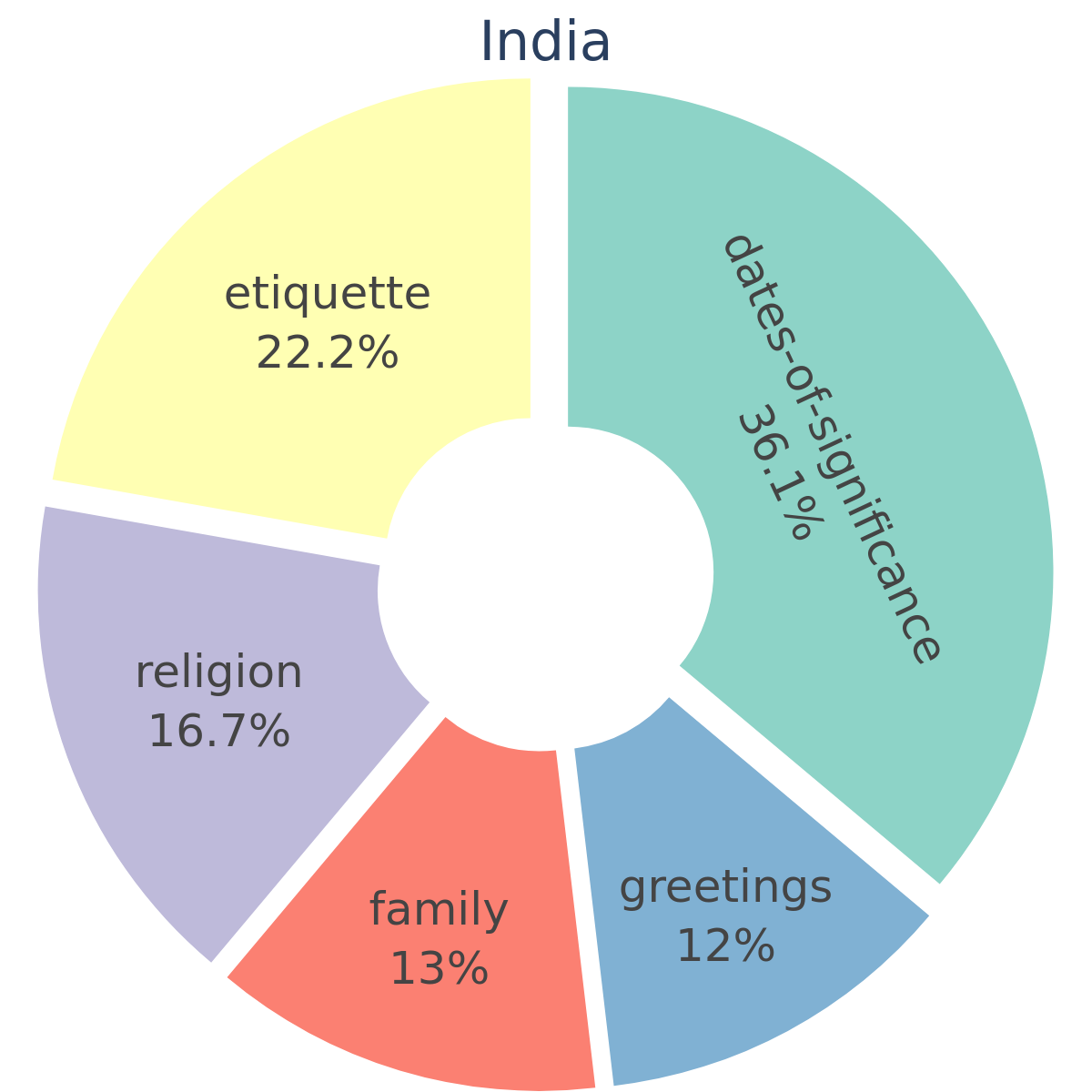}
    \end{subfigure}
    \hfill
    \begin{subfigure}[t]{0.3\textwidth}
        \centering
        \includegraphics[width=\linewidth]{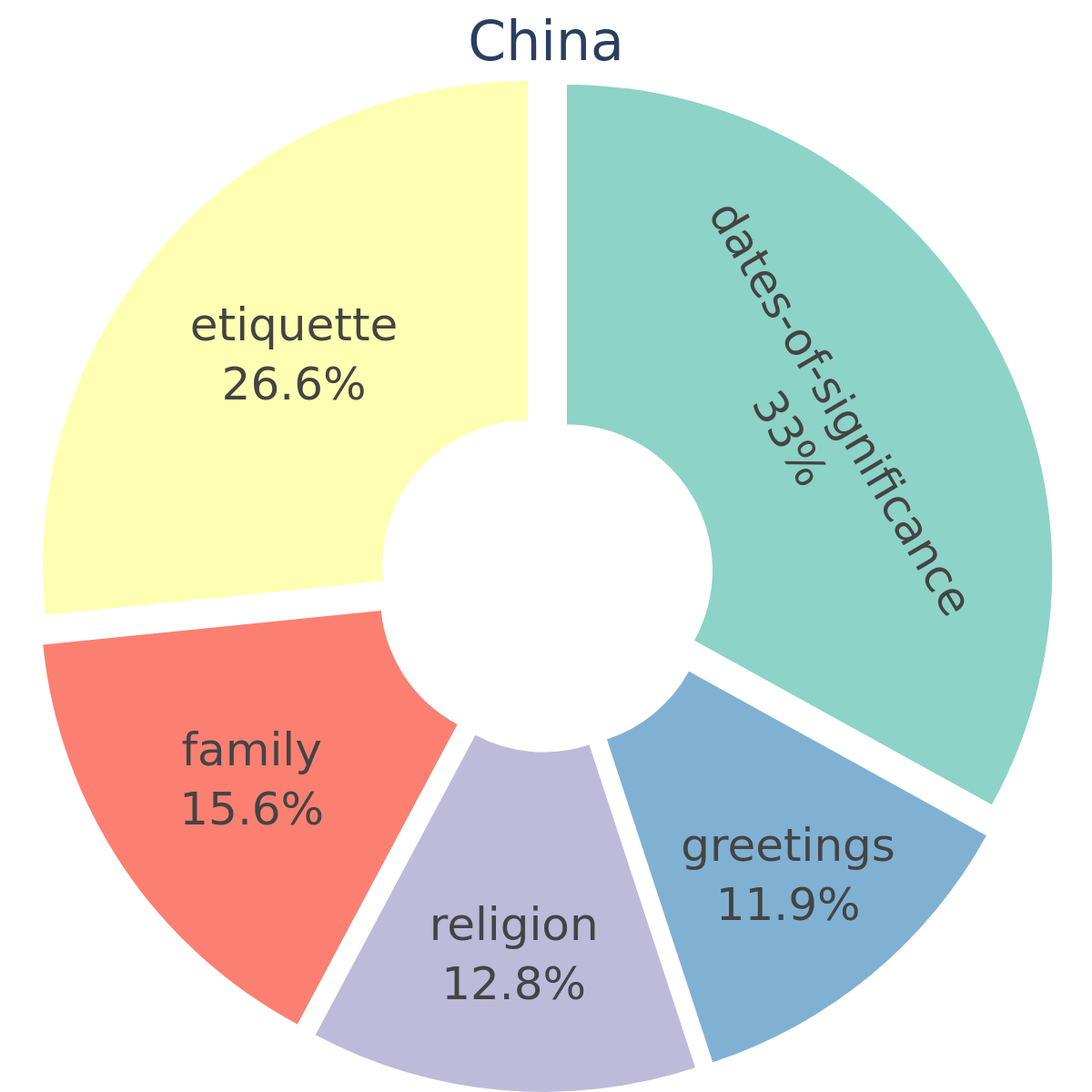}
    \end{subfigure}

    \vspace{1em}
    
    \begin{subfigure}[t]{0.3\textwidth}
        \centering
        \includegraphics[width=\linewidth]{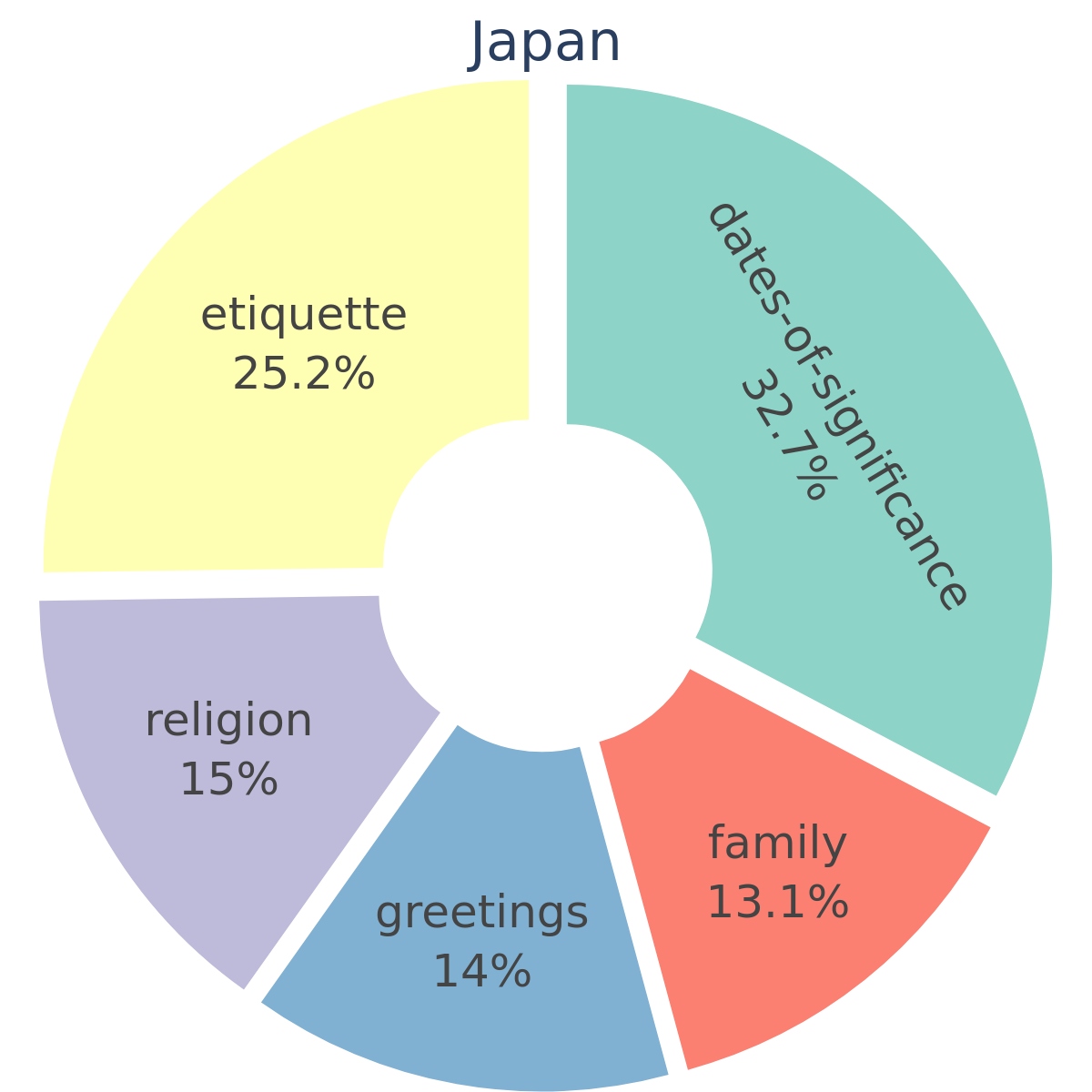}
    \end{subfigure}

    \caption{Distribution of prompts from different categories across countries.}
    \label{fig:plots_grid}
\end{figure*}

\Cref{fig:plots_grid} shows the distribution of prompts across five cultural categories used in constructing \dataset: \textit{dates-of-significance}, \textit{etiquette}, \textit{family}, \textit{religion}, and \textit{greetings}. Across countries, \textit{dates-of-significance} consistently accounts for the largest share of prompts, followed by \textit{etiquette}. This distribution reflects the relative amount of information available for each category in the Cultural Atlas. The remaining three categories—\textit{family}, \textit{religion}, and \textit{greetings}—have relatively balanced proportions. We aimed to maintain a similar category distribution across countries to support fair cross-cultural comparisons. Notably, South Africa lacks sufficient information in the \textit{family} category, so it is excluded from that category in the figure.

\begin{figure*}[ht]
    \includegraphics[width=\textwidth]{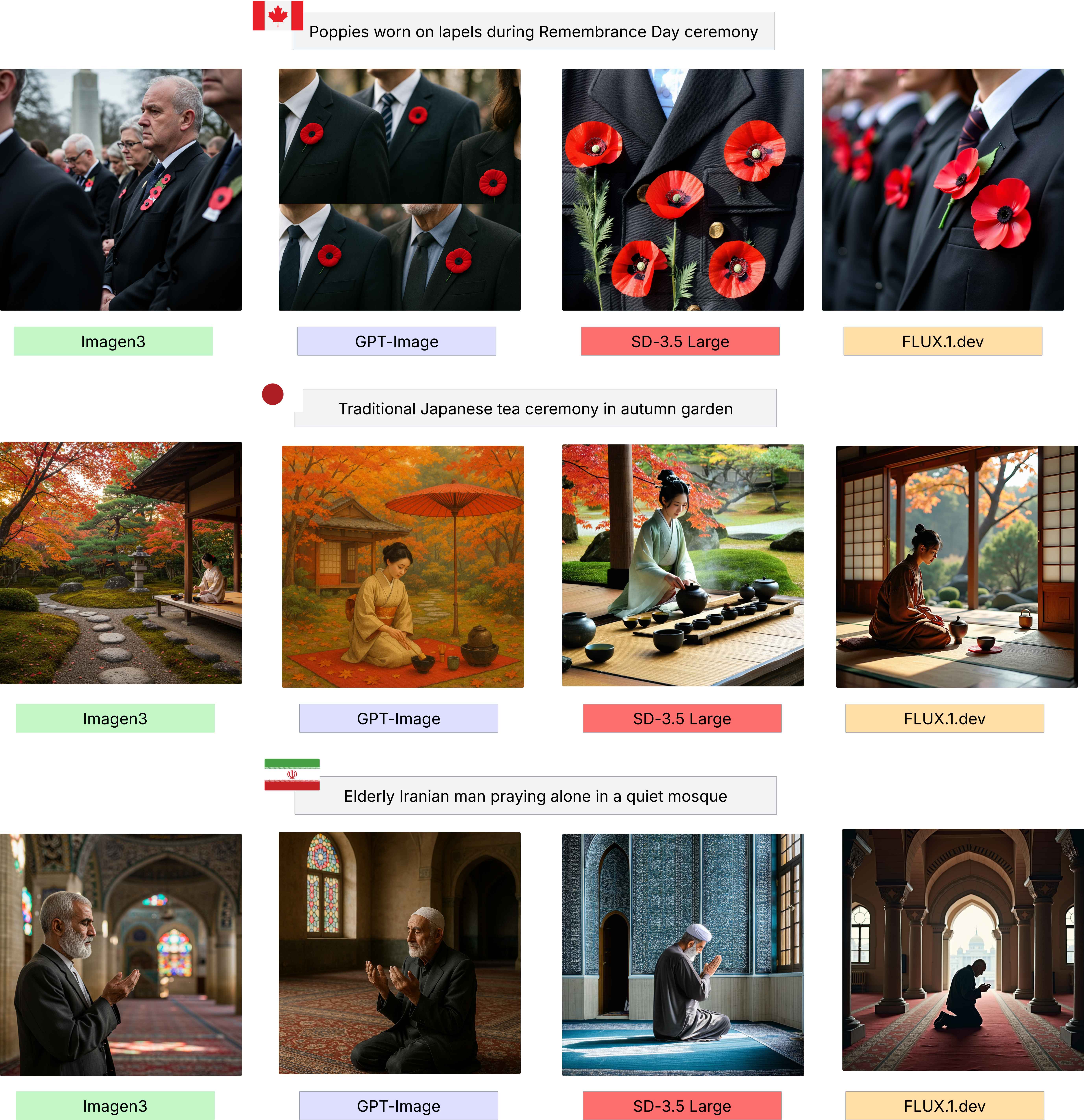}
    \caption{Prompt-image examples from \dataset\ across different countries generated by the models.}
    \label{fig:cultural_frames_examples}
\end{figure*}

\subsection{Image Generation} \label{app:image_generation}

We generate images at a resolution of 1024×1024 across all models to ensure consistency. For GPT-Image, we set the image quality to high. For Imagegen3, we use VertexAI to make API calls and enable the default enhance\_prompt setting, which expands the prompt prior to image generation. For FLUX.1-dev, we set the guidance scale to 3.5, max\_sequence\_length to 512, and use 50 inference steps. In the case of SD-3.5-Large, we use a guidance scale of 4.5 and 40 inference steps. 

\subsection{Prompt-Image Examples}\label{app:prompt_image_examples}
Some examples of prompts along with images generated using different models are provided in \Cref{fig:cultural_frames_examples}.

\subsection{Single Image Generation Analysis}\label{app:single_image}
We generate only one image per prompt due to the practical constraints of our annotation budget and the need to maintain a manageable scale. Despite this limitation, we believe our findings remain meaningful and generalizable, particularly given the known low diversity in model outputs~\citep{kannen2025aestheticsculturalcompetencetexttoimage}. To substantiate this, we conducted two additional analyses:

\paragraph{Quantifying Image Diversity for CulturalFrames} We analyze the diversity of generated images using the best-performing open-source model, Flux.1-Dev~\citep{flux2024}. For every prompt in CulturalFrames, we generate 4 images using different random seeds. We then embed these images using the CLIP model (ViT-L/14@336px)~\citep{radford2021learning} and compute the Vendi Score~\citep{friedman2023vendiscorediversityevaluation}, which reflects the effective number of distinct images in a set. Across all prompts, we find an average Vendi score of 1.5 (standard deviation 0.3) for 4 images, indicating that only 1.5 unique images are produced out of 4 on average. This result confirms the low diversity previously reported in the literature.

\begin{table}[t!]
    \centering
    \begin{tabular}{|p{6cm}|p{7cm}|c|}
        \hline
        \textbf{Prompt} & \textbf{Observed Cultural Issue} & \textbf{Prevalence} \\
        \hline
        Visitors removing shoes before entering a Hindu temple in India. & Annotators commented that people were not removing their shoes, and many were still wearing shoes as they entered the temple. & 4/4 \\
        \hline
        Chinese couple receiving parental blessings in traditional attire. & Annotators observed that there were no parents visible in the images. & 4/4 \\
        \hline
        Families sharing dumplings during Chinese New Year celebration. & Annotators complained that the food shown in the image is ``baozi'' rather than dumplings. & 4/4 \\
        \hline
        Children float Marzanna doll down Polish river to end winter. & Annotators complained that there is no Marzanna doll in the image. & 4/4 \\
        \hline
        Families cooking rice dishes under festive decorations during Pongal. & Annotators pointed out that there was a fire over the rice kept in the dish. & 3/4 \\
        \hline
    \end{tabular}
    \caption{Examples of Persistent Cultural Issues Across Multiple Image Generations}
    \label{tab:cultural_issues_examples}
\end{table}

\paragraph{Checking Generalization of Annotator Comments} To assess whether annotator observations generalize to other images, we manually inspect 4 images each for 20 prompts from India, Poland, and China, countries whose cultural norms our authors are familiar with. These prompts were selected because annotators had already identified cultural issues in the single-image setup.

In all 20 cases, at least three out of four images exhibited the same cultural issues previously flagged. This finding strongly reinforces our initial observations and demonstrates that these issues generalize consistently across multiple generations. \Cref{tab:cultural_issues_examples} provides qualitative examples of prompts and the cultural issues highlighted by annotators, along with the number of images in which these issues were observed.

These results support our claim that even with multiple generations, the same cultural issues tend to persist. This is likely due to the limited diversity of current models. Therefore, while we only use one image per prompt in our main evaluation, our findings do generalize to multi-image settings for current generation systems. Lastly, we believe that the rich explanations collected from annotators can be extremely valuable for future work that studies model biases in multi-image generation settings.

\begin{table}[t!]
    \centering
    \small
    \begin{tabular}{lccccccccccc}
        \toprule
        Gender & Iran & Chile & Germany & Japan & India & China & Canada & South Africa & Brazil & Poland & Average \\
        \midrule
        Male & \textbf{0.68} & \textbf{0.68} & 0.80 & \textbf{0.60} & \textbf{0.80} & \textbf{0.70} & \textbf{0.73} & \textbf{0.84} & 0.82 & 0.74 & 0.74 \\
        Female & \textbf{0.74} & \textbf{0.80} & 0.82 & \textbf{0.53} & \textbf{0.73} & \textbf{0.60} & \textbf{0.80} & \textbf{0.77} & 0.84 & 0.72 & 0.72 \\
        \bottomrule
    \end{tabular}
    \caption{Average image-prompt alignment scores by gender and country. The numbers highlighted have a difference greater than 0.5.}
    \label{tab:gender_agreement}
\end{table}

\begin{table}[t!]
    \centering
    \small
    \begin{tabular}{lccccccccccc}
        \toprule
        Age Group & Germany & Iran & Chile & Japan & India & China & Canada & South Africa & Brazil & Poland & Average \\
        \midrule
        18--24 & \textbf{0.84} & 0.71 & \textbf{0.77} & \textbf{0.69} & 0.74 & \textbf{0.65} & 0.75 & 0.80 & 0.83 & 0.76 & \textbf{0.75} \\
        25--44 & 0.78 & 0.67 & \textbf{0.78} & 0.61 & \textbf{0.78} & \textbf{0.71} & 0.73 & 0.77 & \textbf{0.85} & 0.72 & \textbf{0.74} \\
        45+ & \textbf{0.76} & 0.71 & \textbf{0.45} & \textbf{0.57} & \textbf{0.67} & 0.73 & 0.76 & 0.78 & \textbf{0.77} & 0.72 & \textbf{0.68} \\
        \bottomrule
    \end{tabular}
    \caption{Average image-prompt alignment scores by age groups and country. The numbers highlighted have a difference greater than 0.5.}
    \label{tab:age_agreement}
\end{table}

\subsection{Inter Human Agreement}\label{app:human_agreement}
To establish that our inter-annotator agreement is well within the field's norms, we quantitatively compare our country-level Krippendorff's Alpha and Fleiss' Kappa scores against published values from two closest benchmarks, CUBE~\citep{kannen2025aestheticsculturalcompetencetexttoimage} and CultDiff~\citep{bayramli2025diffusionmodelsgloballens}. For Krippendorff's Alpha, across both image-prompt alignment and image-quality, \dataset's country-level scores consistently match and often exceed the lower bounds of CUBE's reported ranges (e.g., CUBE's image-prompt alignment: 0.09–0.58 vs. \dataset: 0.24–0.42). Similarly, for Fleiss' Kappa, our agreement on prompt alignment (0.179–0.406) and image quality (0.157–0.341) is noticeably higher than CultDiff's general figures (0.07–0.17). For the overall score, where both datasets share a 1–5 scale, our agreement (0.06–0.14) is comparable. Importantly, \dataset\ attains these agreement levels despite requiring raters to judge more subtle, implicit cultural cues than the more object-level signals in the benchmarks. We credit this strong performance to our meticulously designed evaluation framework, which iteratively updated instructions and filtered workers to ensure high data quality.
To understand inter-human agreement for \dataset~ better, we quantitatively and qualitatively analyze several key factors:

\paragraph{Do people of different genders rate images differently?} For every country, we split the annotations by gender and calculate the mean scores provided by each gender for the image-prompt alignment criteria. Our data is predominantly annotated by people who identify as \textit{male} or \textit{female}, except Japan, where 1 annotator did not identify with either gender. Hence, we present the analysis across only these two categories of gender. We make sure to include only those prompt-image instances (2248 of them) where we have ratings from both genders to ensure fair evaluation.

\Cref{tab:gender_agreement} provides the average image-prompt alignment scores provided by male and female annotators. We begin by examining the overall average scores across gender groups: males score 0.74 and females score 0.72, resulting in a modest gap of 0.02. This difference is slightly higher than the 0.01 gap observed when annotations are randomly split, suggesting that gender may play a minor but measurable role in rating variation. However, this effect appears more pronounced when analyzed at the country level.

Several countries in \Cref{tab:gender_agreement} exhibit notable gender-based differences in cultural alignment scores. Chile shows the largest gap, with females scoring 0.80 and males 0.68. China also reflects a considerable difference, with males scoring significantly higher, 0.70 and females 0.60. Canada, India, Japan, and South Africa also demonstrate moderate differences, with females and males differing by over 0.06. These gaps may reflect differences in perception, interpretation, or cultural sensitivity across genders in line with previous works that study gender based variations in T2I evaluation~\citep{rastogi2024insightsdisagreementpatternsmultimodal, rastogi2025viewsafetydeepdive}. Despite these variations, some countries like Germany, Brazil, and Poland show more consistent scores between male and female annotators.

\paragraph{Do people from different age groups rate images differently?}
For each country, we categorize annotators into three age groups (18-24, 25-44, 45+). This corresponds to Gen Z, GenX, and millennials, respectively. We make sure to include only those prompt-image instances (2407 of them) where we have ratings from two of the three age groups (as ensuring all three age groups annotated an instance filtered a lot of annotations, as we collect only 3 human annotations for a prompt-image pair) to ensure fair evaluation. We calculate the average prompt alignment scores and report them in \Cref{tab:age_agreement}.

The age-wise analysis reveals clear generational differences in how cultural alignment is rated. On average, annotators aged 18–24 give the highest scores (0.75), followed closely by the 25–44 group (0.74), while the 45+ group gives notably lower scores (0.68). This 0.07 drop between the youngest and oldest age groups is substantially higher than the 0.01 difference observed when annotations are randomly split (3-way random split, each pairwise difference was ~0.01) and differences are calculated, suggesting that age meaningfully influences evaluation behavior.

On a country level, annotators aged 18–24 assign the highest scores the most number of times (5/10 countries), followed by the 24-44 age group (4/10 countries), suggesting they may be more optimistic, lenient, or culturally flexible. This trend is most prominent in Chile (0.77 for 18-24 vs. 0.45 for 45+), Japan (0.69 for 18-24 vs. 0.57 for 45+), and India (0.78 for 24-44 vs. 0.67 for 45+). In contrast, older participants (45+) tend to give lower scores, indicating more critical assessments, possibly due to deeper cultural anchoring. Countries like Iran, South Africa, and Canada exhibit relatively stable scores across age groups, suggesting less generational variance in perception. This analysis underscores the importance of considering age-based diversity when evaluating subjective alignment tasks, as perspectives can shift meaningfully across generations.

\begin{table*}[h!]
    \centering
    \begin{tabularx}{\textwidth}{XXcXc}
        \toprule
        \textbf{Prompt} & \textbf{Annotator 1 Comment} & \textbf{Score} & \textbf{Annotator 2 Comment} & \textbf{Score} \\
        \midrule
        Chinese villagers gathering for Laba Festival porridge feast & It is not Laba-style porridge & 0 & That is not Laba porridge. Laba porridge contains at least 8 ingredients. & 0.5 \\
        \midrule
        Casual hug between German friends at a cafe & I cannot see a hug. Beer is not typical for a cafe. & 0 & The hug is hard to see. The scene doesn't look casual. The person is drinking beer in a cafe. & 0.5 \\
        \midrule
        Polish family passing pierogi platter during afternoon meal & These are not pierogi. & 0 & These are not traditional Polish dumplings & 0.5 \\
        \midrule
        Canadians lining up outside mall for Boxing Day sales & There isn't much of a line, nor does this suggest the people here are Canadian/in Canada & 0.5 & There is no explicit depiction of Canadians, nor is there a line present, nor is the ``line'' outside the mall. & 0 \\
        \bottomrule
    \end{tabularx}
    \caption{Qualitative examples of different sensitivities in scores shown by annotators for the same or similar issues. The score to the right of the annotator comment is the rating the annotator provided.}
    \label{tab:annotator_discrepancies}
\end{table*}

\paragraph{Are people's sensitivities to the same issues different?}
We analyze whether annotators may provide similar reasoning for their judgments but assign different alignment scores, indicating varying sensitivities to the same issue. We observe such instances in our dataset and argue that this variation is not annotator noise, but a natural outcome of subjective interpretation in value-centric evaluations. The rationales we collect alongside each score are critical in making sense of these differences, offering insight into annotators' thought processes and allowing us to study the nuances behind disagreement, rather than dismissing them as inconsistencies. We include qualitative examples below to illustrate this phenomenon in \Cref{tab:annotator_discrepancies}.

\begin{table*}[t!] 
    \centering
    \begin{tabularx}{\textwidth}{XXX}
        \toprule
        \textbf{Prompt} & \textbf{Annotator 1 Comment} & \textbf{Annotator 2 Comment} \\
        \midrule
        Sikh children learning Gurbani in an Indian classroom & Children look more like a foreigner than an Indian sikh. & Girl Sikh children don't wear turbans. \\
        \midrule
        Families sharing fish meal on Good Friday in Brazil & The image does not depict a fish meal like Brazilian people eat it. In fact, the fish looks raw, so it's weird. & I can't see a Brazilian family in this photo; there is nothing that indicates it. It looks more like Asian people. \\
        \midrule
        Traditional African ceremony in KwaZulu Natal province & Men aren't wearing the traditional dress, which would include animal hide. This is an important part of Zulu culture and wouldn't be changed. & There is nothing resembling KwaZulu Natal province, including the clothing and the scenery. \\
        \bottomrule
    \end{tabularx}
    \caption{Qualitative examples of different annotators providing different reasons for their ratings.}
    \label{tab:further_annotator_discrepancies}
\end{table*}

\paragraph{Do people flag different issues for the same image?}
We observe that in a small number of cases, different annotators identify different issues in the same image, which can stem from their diverse cultural backgrounds and lived experiences. What one annotator flags as a misrepresentation may not even register to another, highlighting the subjectivity inherent to cultural evaluation, which could result in different scores. We provide qualitative examples to illustrate this phenomenon in \Cref{tab:further_annotator_discrepancies}. Further, we note that the combination of diverse perspectives provided by the annotators in these cases collectively covers a broad spectrum of potential issues, leading to a more holistic and robust understanding of cultural expectations. 

\section{Image Rating}\label{app:image_rating}

\begin{figure*}[ht]
    \includegraphics[width=\textwidth]{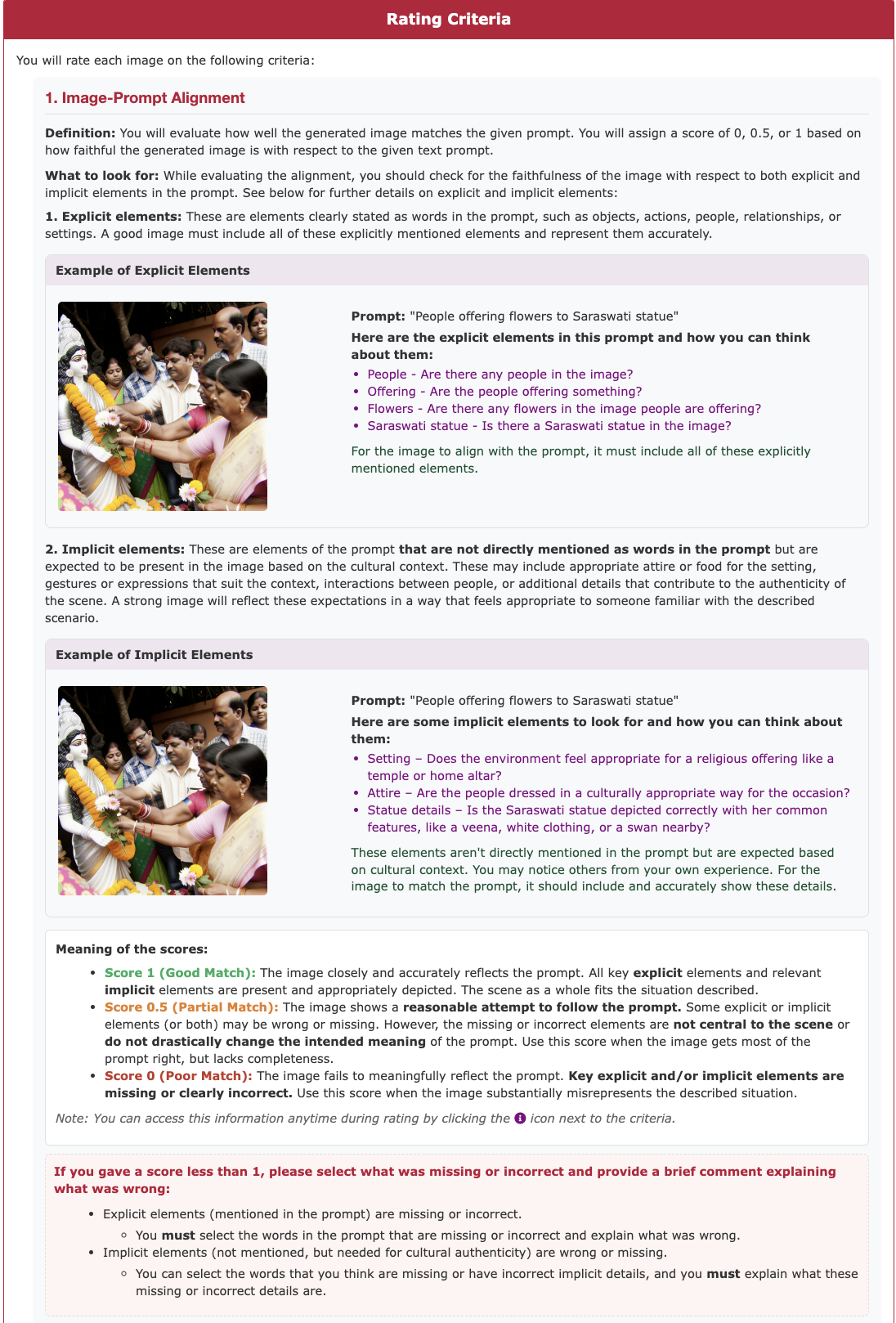}
    \caption{Prompt alignment instructions provided to the annotators. The example shown varies depending on the countries.}
    \label{fig:inst_prompt_alignment}
\end{figure*}

\begin{figure*}[ht]
    \includegraphics[width=\textwidth]{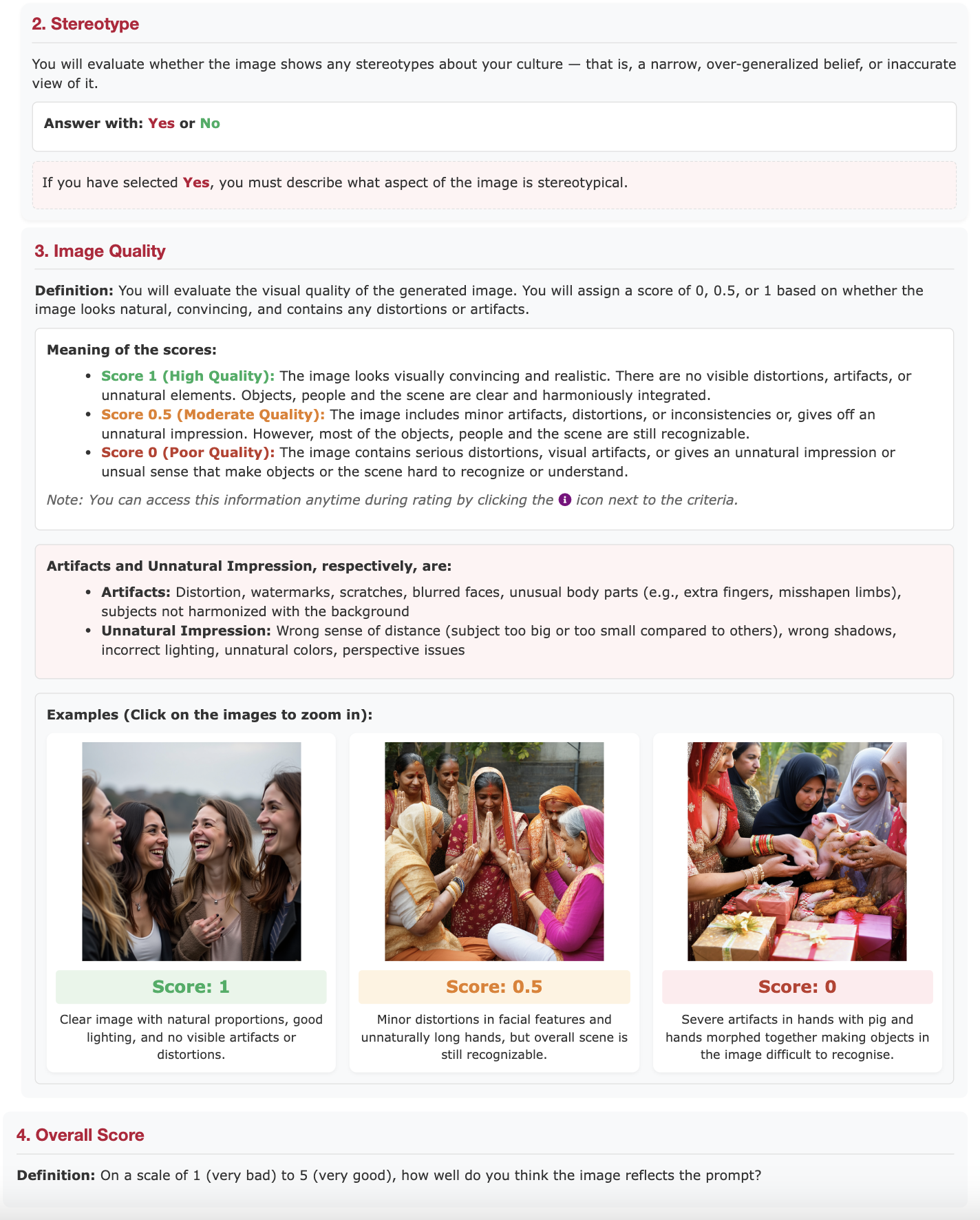}
    \caption{Instructions given to annotators for stereotype, image quality, and overall score criteria.}
    \label{fig:other_instructions}
\end{figure*}

\begin{figure*}[ht]
    \includegraphics[width=\textwidth]{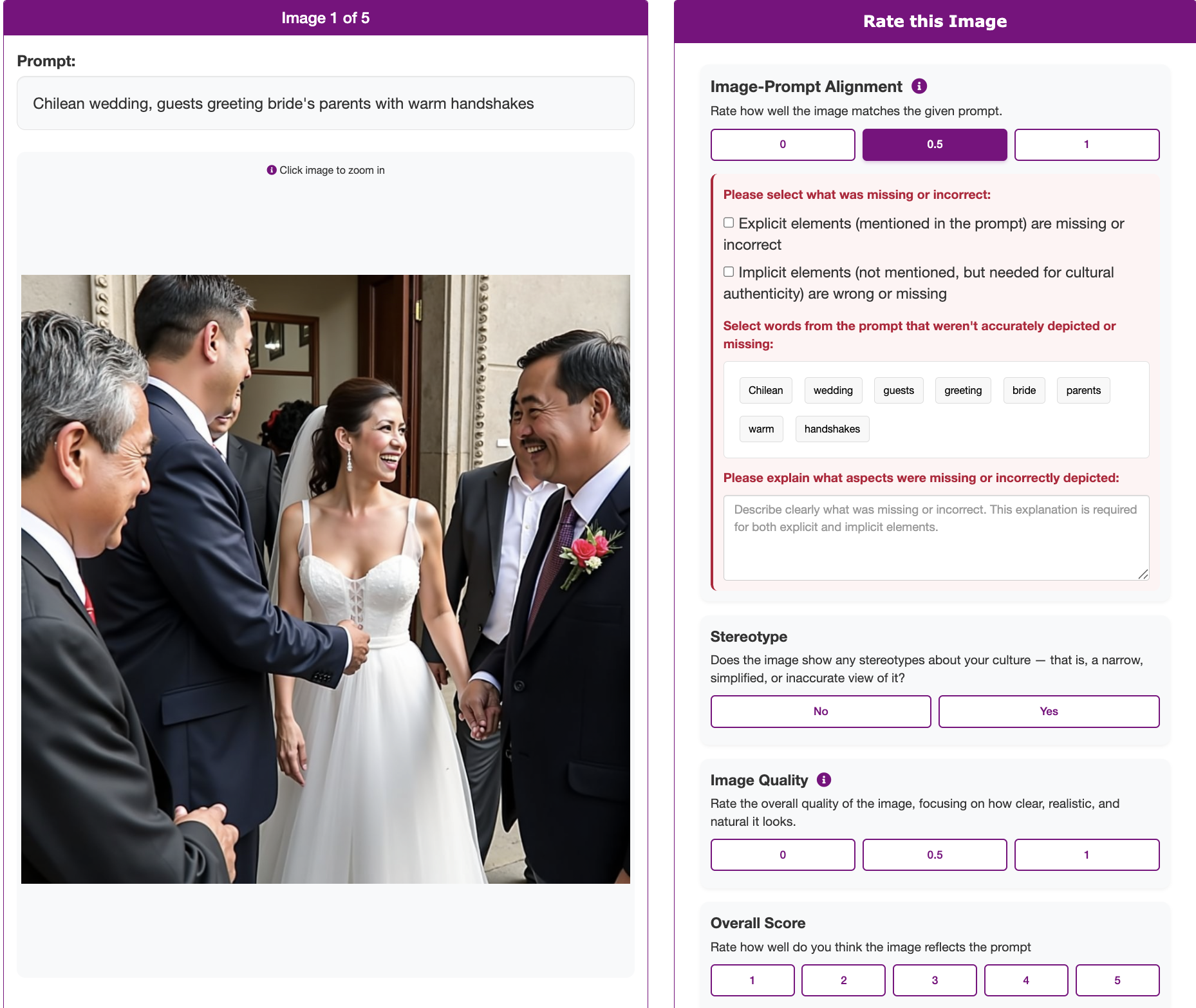}
    \caption{Rating collection interface shown to the annotators. When annotators select a score of less than 1, they need to give detailed feedback regarding explicit and implicit expectations, along with selecting the problematic words.}
    \label{fig:image_rating}
\end{figure*}

\subsection{Rating Interface}
\label{app:ratings}
We develop a custom interface for collecting image ratings. \Cref{fig:inst_prompt_alignment} and \Cref{fig:other_instructions} show the detailed instructions we provide to the annotators for rating images. \Cref{fig:image_rating} shows the interface where annotators rate images. 

\subsection{Annotator Demographics}
\label{app:annotator_demographics}

\Cref{tab:demographics_summary} provides details on the annotators who participated in our studies.

\section{Text-to-Image Models' Analysis}

\subsection{Prompt Expansion Case Study} \label{app:prompt_expansion}

Building on the insights gathered from our detailed analysis of model failures, we propose a simple but effective prompt expansion strategy. Our annotator rationales revealed recurring patterns in what models tend to overlook, such as missing cultural objects, family members, inaccuracies in settings, and mood. To test whether explicitly including these overlooked details in the prompt improves generation authenticity, we selected the 20 lowest-scoring prompts from each country (200 prompts in total across 10 countries) and expanded the prompts using an LLM (Gemini-2.5-Flash). The LLM was given the instructions detailed in \Cref{fig:prompt_expansion}. 

\begin{figure*}[t]
\centering
\small 
\begin{tcolorbox}[title=Prompt Expansion Instructions, myboxstyle]

\textbf{Purpose:}

You are an expert in cultural nuance and creative image generation. Your task is to expand the following brief cultural prompt into a more detailed and descriptive one suitable for a state-of-the-art AI image generator. The goal is to create a visually rich and culturally authentic image.\\

Original Prompt: ``{original\_prompt}''\\

Instructions for Expansion: Enrich the prompt by adding vivid details across these categories:

\begin{itemize}
    \item Setting and Environment: Describe the specific location, time of day, lighting, and background elements. 
    \item People and Demographics: Detail the family members' approximate ages, their relationships to one another, their attire, and their expressions. 
    \item Objects and Food: Specify the types of food on the table, the serving dishes, and any other relevant objects in the scene. 
    \item Cultural Atmosphere and Mood: Capture the overall feeling of the scene—is it lively, warm, formal, or relaxed?
    \item Artistic Style: Suggest a photographic style (e.g., ``cinematic, warm lighting, shallow depth of field, 35mm film look'').
\end{itemize}

Combine all of the above details into a single, cohesive, descriptive paragraph.\\

The output should be in the following format:

Expanded: <expanded\_prompt>

\end{tcolorbox}
\caption{Instructions provided to a LLM to generate expanded prompts.}
\label{fig:prompt_expansion}
\end{figure*}

We generate images using the Flux model (the best open-source model) for these expanded prompts, and use VIEScore to measure the image-prompt alignment accuracy. We use VIEScore as it is the metric that correlates the most with human judgements. We see that there is a consistent improvement of VIEScore (overall score) from 7.3 to 8.4 upon prompt expansion, indicating that careful prompt expansion could indeed help in model improvement.

\begin{figure*}[ht]
    \includegraphics[width=\textwidth]{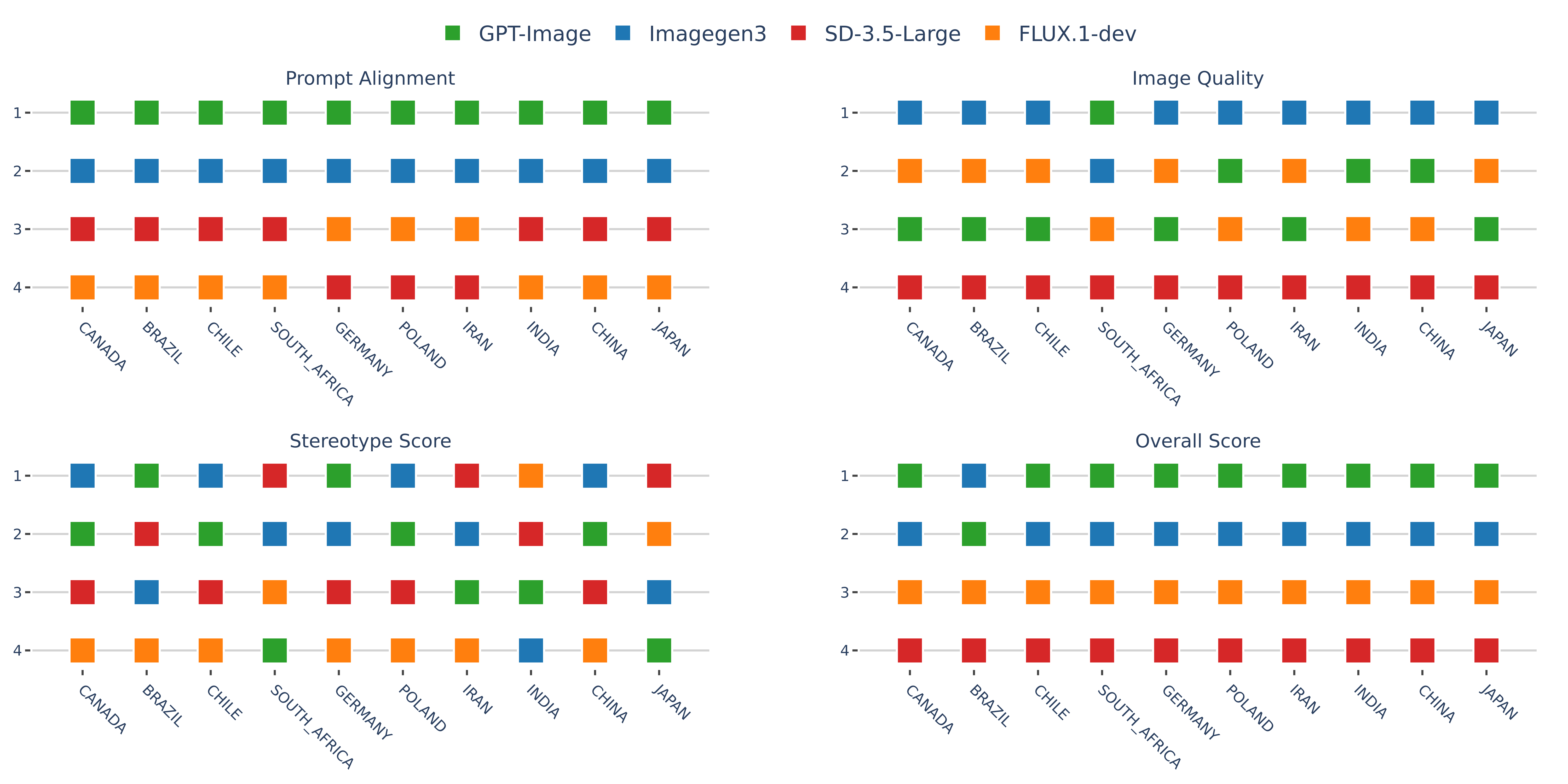}
    \caption{Model ranking across countries for different criteria (1 is the highest rank). Countries are grouped by geographical proximity.}
    \label{fig:model_order}
\end{figure*}

\begin{figure*}[t]
    \centering
    \begin{subfigure}[t]{\textwidth}
        \centering
        \includegraphics[width=\linewidth]{images/models/prompt_alignment_country_analysis.png}
        \caption{Average prompt alignment scores across countries for different models}
        \label{fig:country_analysis_prompt_alignment}
    \end{subfigure}

    \begin{subfigure}[t]{\textwidth}
        \centering
        \includegraphics[width=\linewidth]{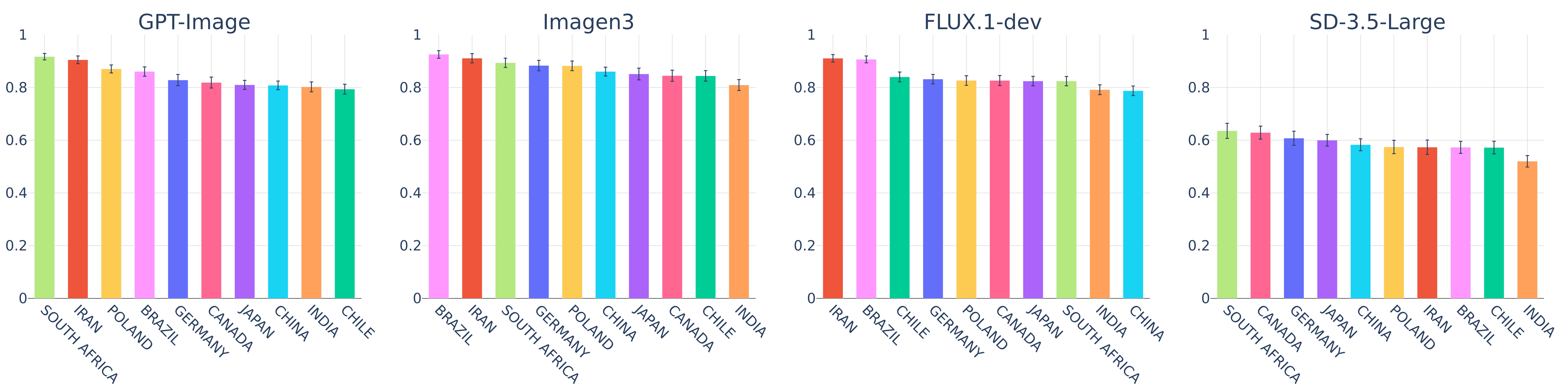}
        \caption{Average image quality scores across countries for different models}
        \label{fig:country_analysis_image_quality}
    \end{subfigure}

    \begin{subfigure}[t]{\textwidth}
        \centering
        \includegraphics[width=\linewidth]{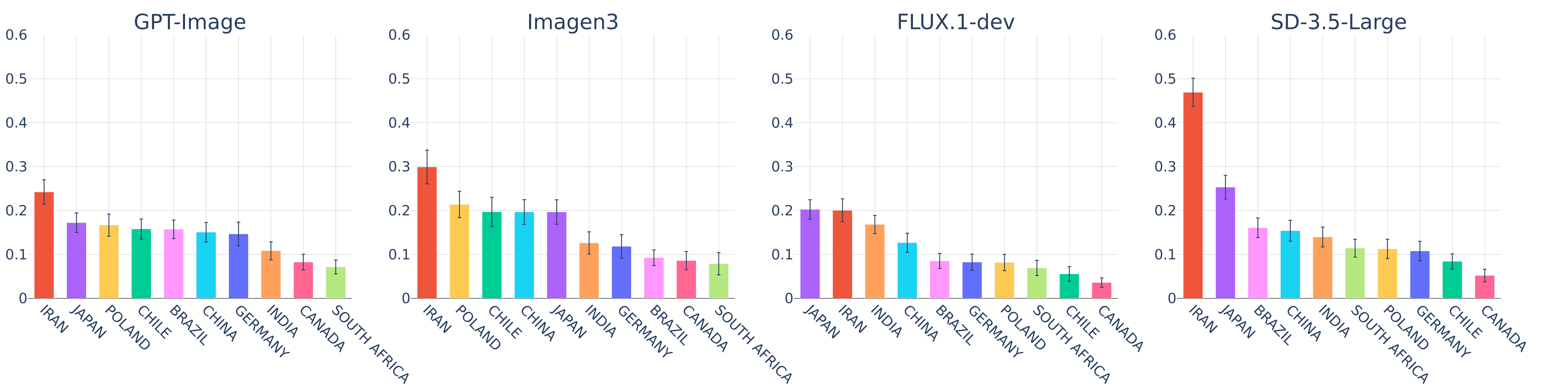}
        \caption{Average stereotype scores across countries for different models}
        \label{fig:country_analysis_stereotype}
    \end{subfigure}

    \begin{subfigure}[t]{\textwidth}
        \centering
        \includegraphics[width=\linewidth]{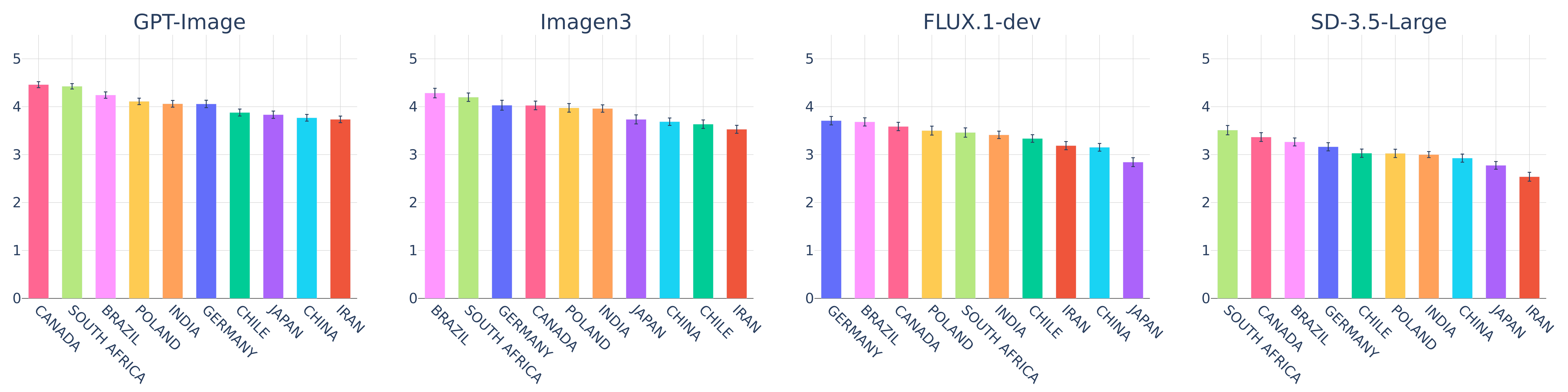}
        \caption{Average overall scores across countries for different models}
        \label{fig:country_analysis_overall_score}
    \end{subfigure}
    \caption{Comparison of different models' scores for different countries for prompt-alignment, image quality, stereotypes, and overall score.}
    \label{fig:all_criteria_model_comparision}
\end{figure*}

\begin{figure*}[t]
    \centering
    \includegraphics[width=1\textwidth]{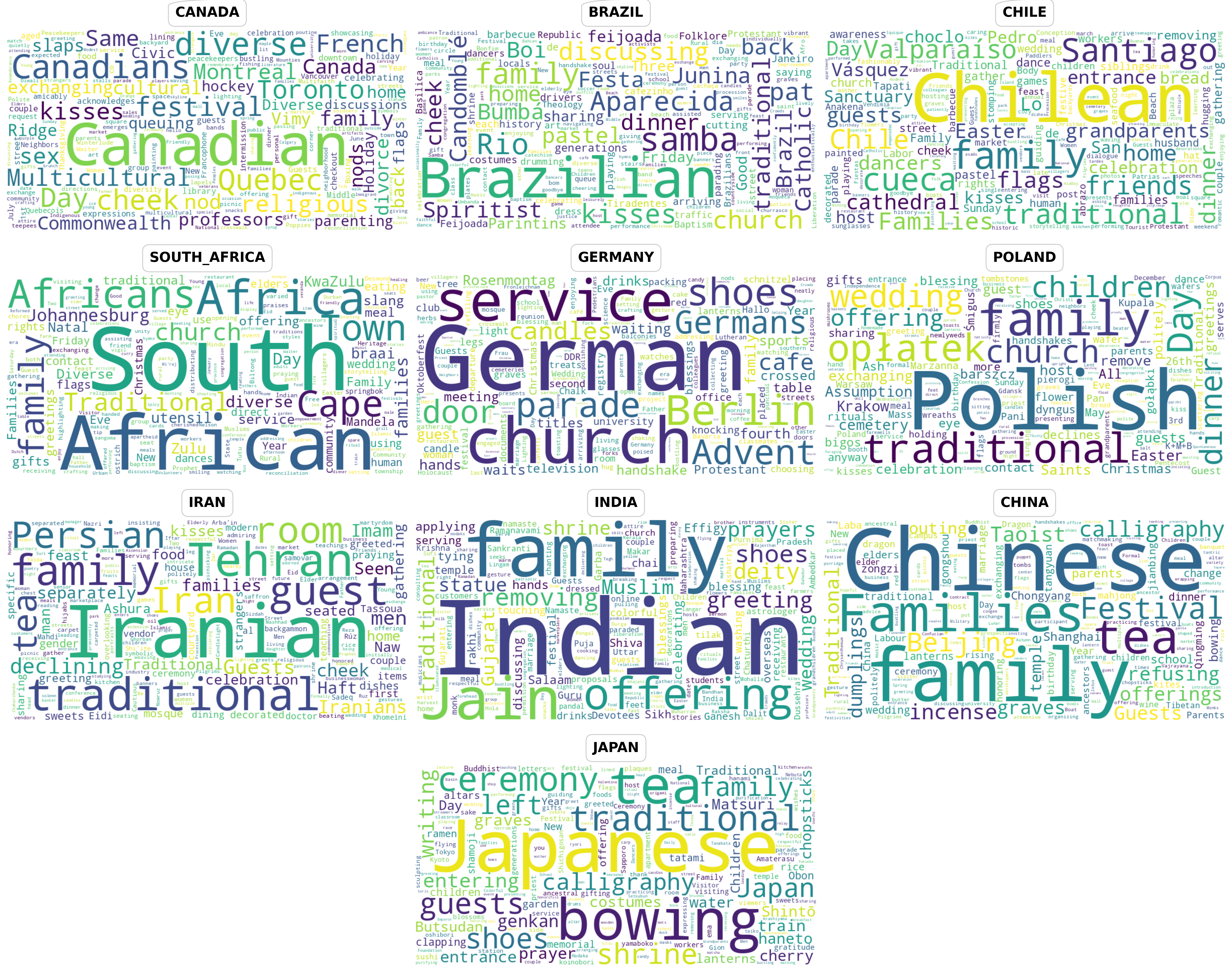}
    \caption{World cloud for words highlighted as having issues by annotators across different countries.}
    \label{fig:wordcloud}
\end{figure*}

\begin{figure*}[t]
\centering
\small 
\begin{tcolorbox}[title=LLM-as-Judge Evaluation Instructions, myboxstyle]

You are a strict yet fair evaluator. You will be given a prompt, issues highlighted by several annotators along with the words which have the issues as marked by the annotators, and an explanation of the automatic metric for how good the image is.

Your task is to assess how well the automatic explanation captures the concerns raised by the annotators.\\

\textbf{TASK}
\begin{itemize}
    \item \texttt{ORIGINAL\_PROMPT} – the text that generated the image
    \item Up to four annotator blocks, each with:
    \begin{itemize}
        \item \texttt{HUMAN\_REASON\_X} – A 1-2 sentence critique
        \item \texttt{HIGHLIGHTED\_WORDS\_X} – Prompt words flagged by that annotator
    \end{itemize}
    \item \texttt{MODEL\_REASON} – The automatic explanation
\end{itemize}

Decide how well \texttt{MODEL\_REASON} covers the \textbf{union} of concerns across all annotators.\\

\textbf{Coverage Scale}
\begin{itemize}
    \item \textbf{5 (Perfect)} – Covers all issues highlighted by annotators with no contradictions.
    \item \textbf{4 (Strong)} – Covers most main concerns, may miss at most one minor issue.
    \item \textbf{3 (Partial)} – Covers around half of the union of concerns.
    \item \textbf{2 (Weak)} – Only covers a small portion; many key points are missing or vague.
    \item \textbf{1 (None/Wrong)} – Irrelevant explanation or contradicts annotators.
\end{itemize}

\textbf{Output Format}
\begin{verbatim}
{
    ``score'': 1-5,
    ``explanation'': ``1-2 sentence explanation of the score''
}
\end{verbatim}

\textbf{Rules}
\begin{itemize}
    \item Sometimes, annotators highlight specific words without explicitly explaining them in their comments. In such cases, it should be assumed that these words indicate an issue, and the metric explanation should mention that these words have issues.
    \item If \texttt{MODEL\_REASON} contradicts the general consensus of the annotators, assign a score of 1.
    \item Mention missing or covered ideas in no more than 50 words.
    \item Output \textbf{only} a valid JSON object as shown above.
\end{itemize}

\end{tcolorbox}
\caption{Instructions for LLM-as-a-judge evaluation to assess the alignment between VIEScore's reasoning and human concerns on a 1–5 Likert scale.}
\label{fig:llm_judge}
\end{figure*}

\begin{table}[t]
\centering

\caseRow{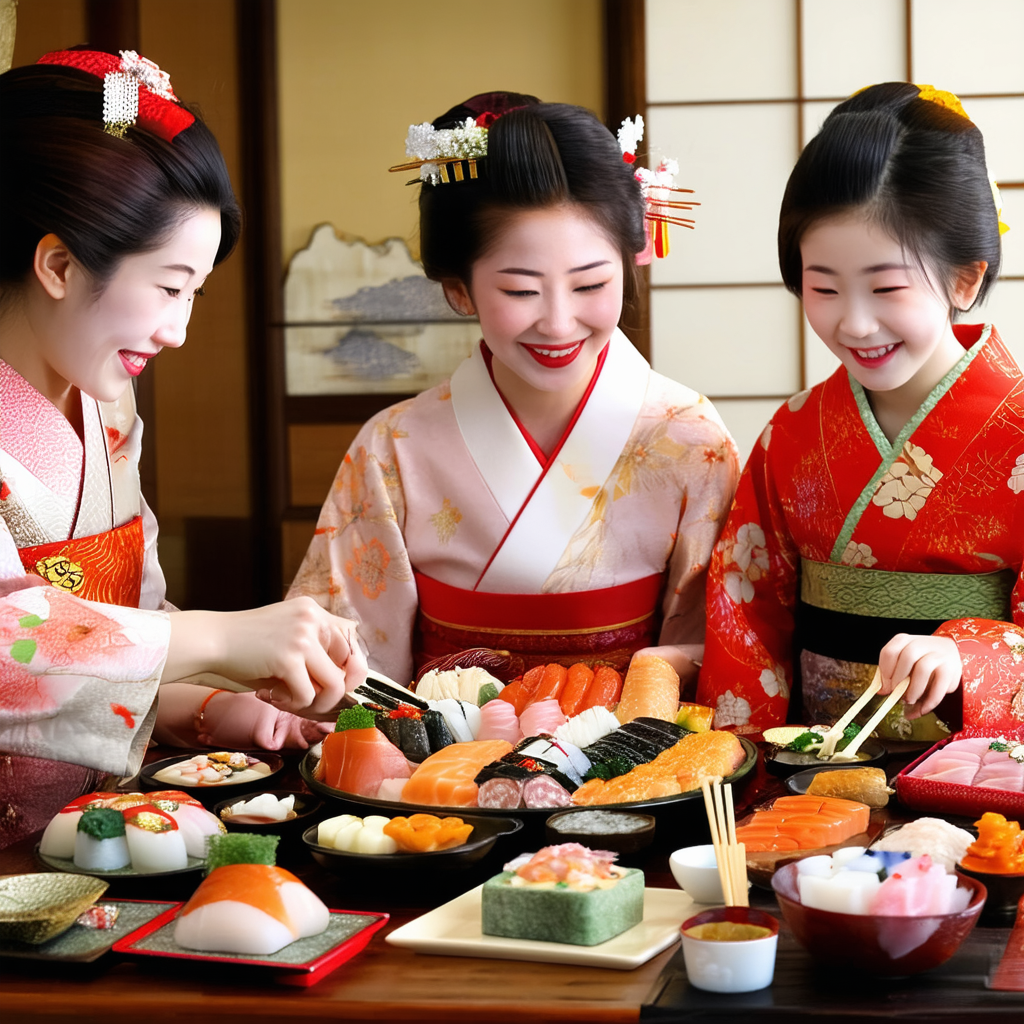}
{Japanese family celebrating New Year's at home with osechi-ryori}
{Usually, the Japanese don't make their face white on New Year's. The way of holding chopsticks is strange. The foods look like suhi rather than osechi ryori.}
{}
{The image accurately reflects the prompt, showing a Japanese family in traditional kimonos celebrating New Year. Osechi-ryori is prominently featured, and the home setting with sliding doors matches cultural expectations. Implicit cues like traditional attire, joyful expressions, and a family-oriented atmosphere further enhance authenticity.}
{}

\caseRow{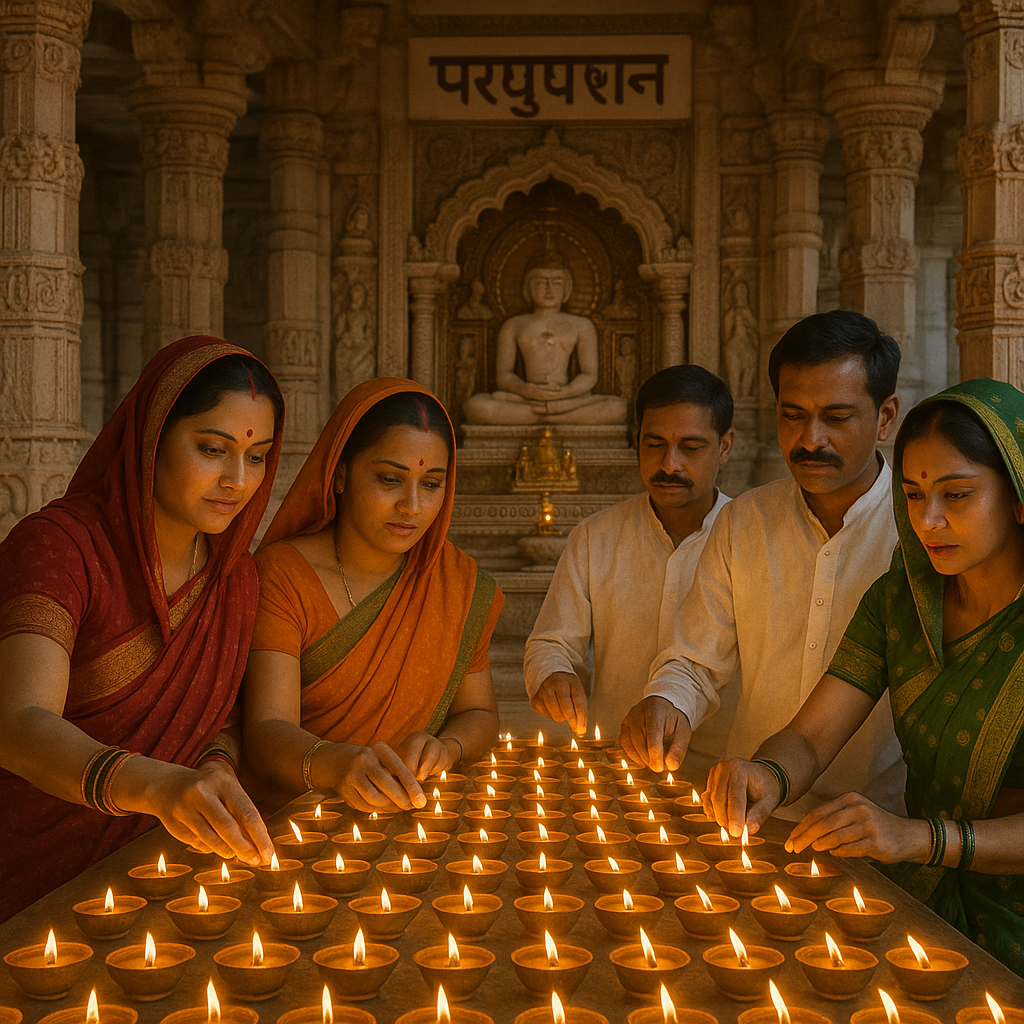}
{Devotees lighting candles at a Jain temple during Paryushana in India.}
{Paryushana would not be engraved inside the temple. There are no candles, and how can they light them with their hands?}
{}
{The image effectively depicts devotees lighting candles in what appears to be a Jain temple setting. The architecture and attire of the individuals are consistent with an Indian cultural setting, and the script in the background corresponds to the theme of Paryushana. The ambiance and lighting convey a religious and serene atmosphere. The missing point could be a more explicit reference to Paryushana elements.}
{}

\caseRow{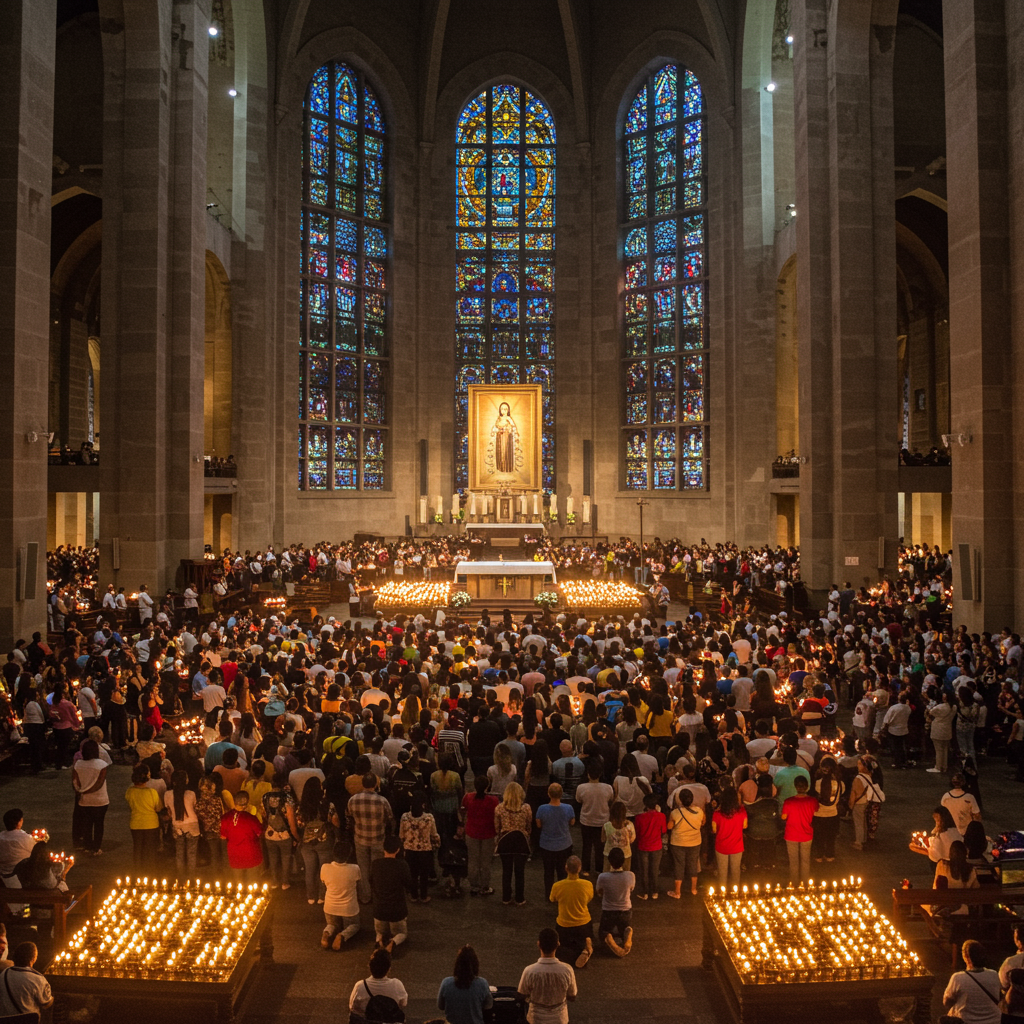}
{Faithful lighting candles at Aparecida shrine in Brazil}
{The church shown in the generated image doesn't look like the Aparecida Shrine but a generic catholic church.}
{}
{The image captures a large number of people in a religious setting, with many candles lit, fitting the description of faithful lighting candles. The architecture resembles a shrine, and the stained glass windows add to the spiritual ambiance. The scene closely aligns with the Aparecida shrine in Brazil. The only minor point could be about distinct visual cues specific to Brazil, which might not be strongly evident.}
{}

\caption{Examples showing mismatches between human and VIEScore rationales.}
\label{tab:human-metric-rationale}
\end{table}

\begin{figure*}[t]
\centering
\small 
\begin{tcolorbox}[title=Explicit instructions provided to VIEScore, myboxstyle]

You are a professional cultural visual analyst and digital artist. You are tasked with evaluating how faithfully an AI-generated image represents the provided text prompt, especially in capturing both explicit and culturally implicit details.\\

All input images are AI-generated. All people or characters shown are also AI-generated — you do not need to consider privacy or ethical issues in the evaluation.\\

Your evaluation should result in a JSON output in the following format:\\

\begin{verbatim}
{
    ``reasoning'': ``<reasoning to determine how faithful the image is to the prompt>'', 
    ``score'': [<integer from 0 to 10>],
}
\end{verbatim}

How to Evaluate:\\

You will give a score from 0 to 10, based on how accurately the image matches the explicit and implicit elements described in the prompt.

1. Explicit Elements: 
Explicit elements are the clearly stated words in the prompt — such as objects, people, actions, locations, or relationships. A good image must include and visually represent all of these elements clearly and correctly.

\begin{verbatim}
{country specific example}
\end{verbatim}

You should check: Are all these elements present and recognizable? Is their interaction depicted as described?\\

2. Implicit Elements
These are elements of the prompt that are not directly mentioned as words in the prompt but are expected to be present in the image based on the cultural context. These may include appropriate attire or food for the setting, gestures or expressions that suit the context, interactions between people, or additional details that contribute to the authenticity of the scene. A strong image will reflect these expectations in a way that feels appropriate to someone familiar with the described scenario.\\

For the same prompt above, implicit elements may include:

\begin{verbatim}
{country specific example}
\end{verbatim}

There may be several other implicit details that need to be considered given the image and the prompt. For the image to align with the prompt, it should include and accurately show these details.\\

From scale 0 to 10: \\
A score from 0 to 10 will be given based on the success in following the prompt. \\
(0 indicates that the AI-generated image does not follow the prompt at all, and major explicit elements and implicit elements are missing or incorrectly depicted. 10 indicates the AI-generated image follows the prompt perfectly, and all explicit elements and necessary implicit elements are present and correctly depicted.)\\

Put the score in a list such that output score = [score].\\

Text Prompt: <prompt>

\end{tcolorbox}
\caption{Updated instructions provided to VIEScore, similar to those human raters use to judge images.}
\label{fig:new_viescore}
\end{figure*}

\end{document}